\documentclass[lettersize,journal]{IEEEtran}
\usepackage{amsmath,amsfonts}
\usepackage{algorithmic}
\usepackage{algorithm}
\usepackage{array}
\usepackage{tabularx}
\usepackage{makecell}
\usepackage{colortbl}
\definecolor{mypink}{rgb}{0.9686,0.651,0.694}
\definecolor{mygreen}{rgb}{0.698,0.8588,0.7255}
\definecolor{myblue}{rgb}{0.7216,0.898,0.9804}
\usepackage{textcomp}
\usepackage{stfloats}
\usepackage{url}
\usepackage{verbatim}
\usepackage{stfloats}
\usepackage{graphicx}
\usepackage{float}
\usepackage{cite}
\usepackage{booktabs}
\usepackage{multicol}
\usepackage{multirow}
\usepackage{subcaption}
\usepackage[colorlinks,
            linkcolor=blue,
            anchorcolor=blue,
            citecolor=blue]{hyperref}
\begin{document}

\title{Rethinking the Key Factors for the Generalization of Remote Sensing Stereo Matching Networks}

\author{Liting Jiang,~\IEEEmembership{Graduate Student Member,~IEEE,} Feng Wang, Wenyi Zhang, Peifeng Li, \\Hongjian You, Yuming Xiang,~\IEEEmembership{Senior Member,~IEEE}
\thanks{L.~Jiang, F.~Wang, W.~Zhang, P.~Li, H.~You, Y.~Xiang are with Aerospace Information Research Institute, Chinese Academy of Sciences, Beijing 100094, China; they are also with the Key Laboratory of Technology in Geo-spatial Information Processing and Application System, Chinese Academy of Sciences, Beijing 100190, China and Key Laboratory of Target Cognition and Application Technology, Chinese Academy of Sciences, Beijing 100190, China. L.~Jiang, H.~You, Y.~Xiang are also with School of Electronic, Electrical and Communication Engineering, University of Chinese Academy of Sciences, Beijing 100049, China.}
\thanks{This work was supported in part by Key Research Program of Frontier Sciences, Chinese Academy of Sciences, under Grant ZDBS-LY-JSC036 and in part by the National Natural Science Foundation of China under Grant 61901439. (\textit{Corresponding author:Yuming Xiang.})}}

\markboth{Journal of \LaTeX\ Class Files,~Vol.~14, No.~8, August~2021}%
{Shell \MakeLowercase{\textit{et al.}}: A Sample Article Using IEEEtran.cls for IEEE Journals}


\maketitle
\vspace{-10pt}
\begin{abstract}
Stereo matching, a critical step of 3D reconstruction, has fully shifted towards deep learning due to its strong feature representation of remote sensing images. However, ground truth for stereo matching task relies on expensive airborne LiDAR data, thus making it difficult to obtain enough samples for supervised learning. To improve the generalization ability of stereo matching networks on cross-domain data from different sensors and scenarios, in this paper, we dedicate to study key training factors from three perspectives. (1) For the selection of training dataset, it is important to select data with similar regional target distribution as the test set instead of utilizing data from the same sensor. (2) For model structure, cascaded structure that flexibly adapts to different sizes of features is preferred. (3) For training manner, unsupervised methods generalize better than supervised methods, and we design an unsupervised early-stop strategy to help retain the best model with pre-trained weights as the basis. Extensive experiments are conducted to support the previous findings, on the basis of which we present an unsupervised stereo matching network with good generalization performance. We release the source code and the datasets at \url{https://github.com/Elenairene/RKF\_RSSM} to reproduce the results and encourage future work.
\end{abstract}
\vspace{-5pt}
\vspace{-7pt}
\section{Introduction}
\IEEEPARstart{R}{emote} sensing 3D reconstruction techniques utilize satellite images to construct Digital Surface Model(DSM) to provide a 3D perspective for geographical analysis. DSM has promising applications in a wide range of fields, such as terrain analysis, disaster monitoring, and digital cities, thus receiving growing attention. Stereo matching is the most critical step in 3D reconstruction, which produces pixel-to-pixel matching for stereo pairs to obtain the disparity map. By combining the camera model with stereo matching method, the position of every point in 3D space can be obtained, which determines the accuracy and quality of 3D reconstruction.

Supervised stereo matching methods rely on the disparity ground truth of stereo pairs to provide supervision information, where methods that generate disparity ground truth by LiDAR (e.g., the KITTI dataset~\cite{KITTI}) are costly and require manual labelling to correct errors caused by reflectance values and occlusions. Methods that obtain disparity ground truth through structured light sensors and mannual labelling (e.g., the MiddleBury dataset~\cite{MiddleBury}), although less expensive, require extensive texture of imaged areas and are only suitable for indoor task. Building synthetic datasets (e.g., SceneFlow~\cite{sceneflow}) is not bothered by the above two problems, but have a gap with the real-world data, bringing difficulties in generalising to real scenes. Therefore, obtaining a large number of real-world stereo pairs with ground truth is not an easy task, and considering growing demands of stereo matching tasks in different scenarios, it is not always feasible to re-obtain the current scenario datasets for training. Previous work has explored unsupervised stereo matching methods, although unsupervised stereo matching datasets are not constrained by ground truth collection, unsupervised models are more likely to be less accurate than supervised models in most task scenarios~\cite{stereoreview}, and stereo pairs from real-world scenarios do not always fully satisfy the assumption of left-right consistency, which is a foundational assumption for unsupervised loss. Thus ideal unsupervised stereo matching datasets are not abundant either.

For remote sensing stereo matching task, the disparity ground truth come from airborne LiDAR point clouds, which are expensive and difficult to synchronise with the satellite image pairs. Aiming at constructing stereo pairs without ground truth, researchers still need to collect and generate stereo pairs from satellite images. However, despite satellite data has been increasing, unlike three-line array stereo mapping satellite images, there is no strict stereo relationship between many satellite images~\cite{rsxuanqi}, which means not all satellite images are suitable for generating stereo pairs, making it difficult for researchers to collect sufficient training data not only for supervised model but also for unsupervised model. Therefore, there is a more urgent need to improve generalization capability of remote sensing stereo matching networks.
\begin{figure}[!htbp]
  \centering
    \includegraphics[width=3.5in]{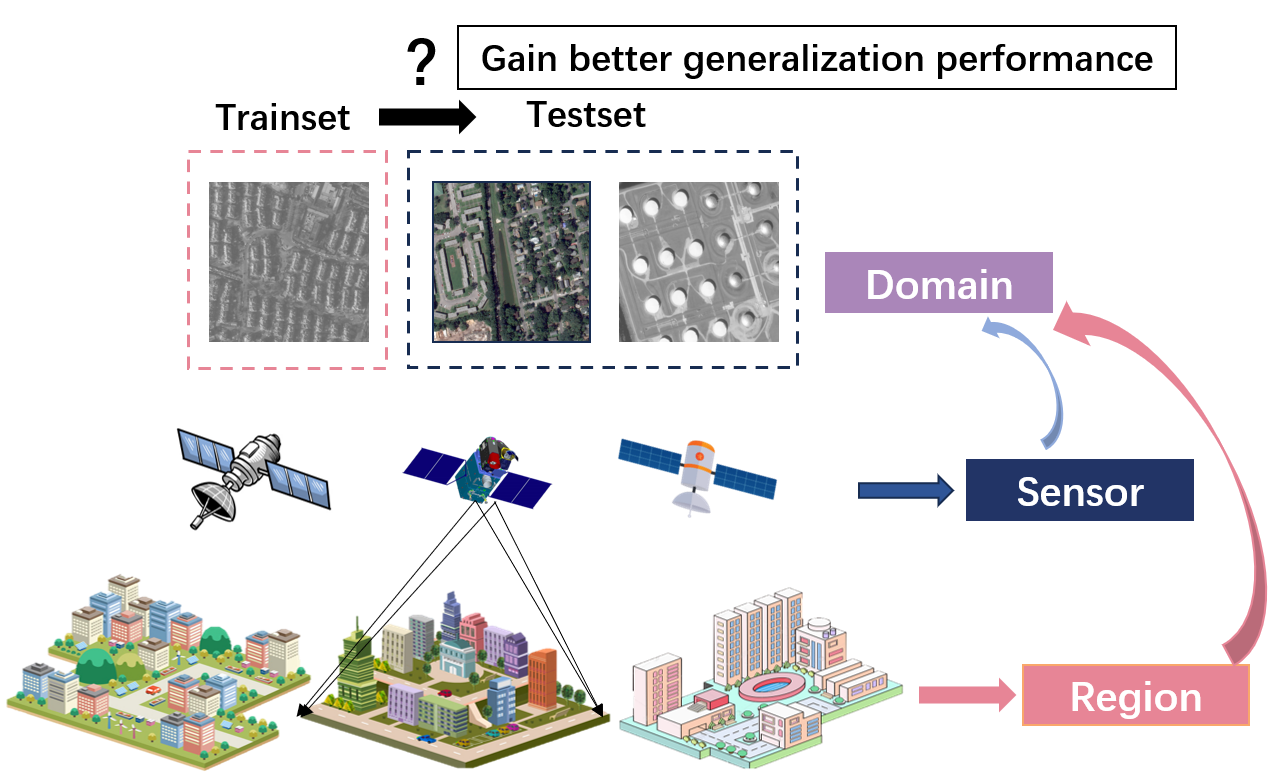}
  \captionsetup{justification=raggedright}
  \caption{Different regional target feature distribution and different sensor will bring different satellite stereo pairs. We define data of the same city and the same sensor as the same domain. The goal is to analyze the key training factors to improve generalization performance for test data from different domains.}
  \label{fig:intro_fig}
  \vspace{-3pt}
\end{figure}

Considering different regional economic levels and different architectural styles, the regional distribution of target appearances varies significantly. In addition, the same target imaged by different sensors may exhibit distinct appearances due to different radiometric and spectral characteristics. Therefore, we define satellite images from the same city and the same sensor as the same domain. As a result, it's rare to have a training set from the exact same domain as the test set, as the concept map shows in Fig.~\ref{fig:intro_fig}. Consequently, enhancing the generalization capability of stereo matching models is a highly desirable goal, not only for the prospective of cost savings in dataset construction but also for saving computational resources and research time. Given the extensive research on stereo matching, studying and demonstrating the factors that are crucial to the model's performance can help researchers around the world to quickly get a model that is suitable for their task. 
While some studies have focused on enhancing the generalization performance of stereo matching networks by proposing new model structure, there is a lack of systematic work on studying what kind of training data, model structure, training manner, and stopping criterion can be used to obtain the expected model that performs better on target data. 

Our work concentrates on remote sensing stereo matching and explores factors that affect the generalization ability of models. It aims to help researchers use limited data to train models that generalize well on images over a wider area and from diverse sensors. Specifically, our contributions are:

1. We systematically analyze multiple factors that may affect the generalization ability of stereo matching network, including the data used for training, network architectures, and different training manners.

2. We propose techniques for model training to improve the generalization ability, such as choosing criteria for training data, early stopping strategy for training iterations, and unsupervised loss design. Accordingly, we present the models with strong generalization ability trained based on the above techniques.

3. We propose a complete procedure to transform any supervised generic stereo matching network to a stereo matching network that conforms to satellite images while applying both supervised and unsupervised training methods.

In the following sections, we begin by summarizing related work in Sec.~\ref{sec:relatedwork}. Subsequently, we elucidate the proposed method for analyzing key factors in Sec.~\ref{sec:methods}. Sec.~\ref{sec:exp} presents the experimental results, illustrating factors influencing the generalization ability of stereo matching network. Finally, we draw conclusions based on the findings of the study in Sec.~\ref{sec:conculsion}.
\vspace{-7pt}
\section{Related Work}
\label{sec:relatedwork}
In order to investigate the key factors for the generalization performance of remote sensing stereo matching, here we briefly summarize work related to this research topic. The related work will be presented in the following order. Firstly we introduce stereo matching network architectures, then we summarize datasets including satellite datasets specially constructed for stereo matching. Finally, we review works dedicated to promote generalization performance of stereo matching.
\vspace{-7pt}
\subsection{Network Architectures}
Based on the powerful feature representation capability of deep learning, neural networks have been widely used in stereo matching tasks. Popular neural network architectures still follow traditional stereo matching framework, containing four steps: feature extraction and fusion, cost computation and aggregation, disparity estimation, and disparity refinement. Zbontar and LeCun et al.~\cite{zbontar} proposed MC-CNN by applying CNN in learning the matching cost between image pairs for the first time. The first end-to-end stereo matching method DispNet was based on 2D convolutions for cost aggregation to compute disparity in 3D cost space~\cite{DispNet}. GCNet was the first stereo matching network that used 3D convolutions, the 3D convolutional cost computation pioneered by this work has been widely adopted ever since~\cite{GCNet}. Landmark work PSMNet proposed a pyramid feature extraction network and used stacked 3D hourglass network for cost aggregation to estimate high-precision disparity~\cite{PSMNet}. The network RAFT introduced a recurrent neural network that iteratively refines the flow predictions, many stereo matching networks based RAFT~\cite{RAFT} appeared afterwards, such as RAFT-Stereo~\cite{raftstereo} and CREStereo~\cite{CREStereo}. Since Google proposed Transformer in 2017, the stereo matching field has also seen the emergence of STTR~\cite{STTR}, a depth estimation algorithm based on Transformer structure, which is advantageous in large untextured regions. Supervised stereo matching methods, despite their excellent performance on several stereo matching datasets, their application requires the datasets construction containing ground truth as supervision. However, capturing real-world disparity ground truth requires structured light (which is suitable only for indoor environments) or LiDAR (which is costly and requires manual error correction). This leads to the fact that the available datasets with ground truth for researchers are restricted and not always feasible to construct for new scenarios. Therefore the generalization of stereo matching networks is important. CFNet~\cite{CFNet} flexibly adapts to stereo matching image pairs of different resolutions by extracting multi-scale features, with has good robustness to different test domains.

Unsupervised methods have been introduced to stereo matching tasks because of the advantage of getting rid of ground truth dependency. Garg et al.~\cite{Garg} first proposed to use reconstruction loss to train the network in a monocular depth estimation task based on the assumption of left-right image luminance consistency. Godard et al.~\cite{Godard} introduced the left-right consistency criterion into stereo matching method to build an unsupervised approach. Most current unsupervised stereo matching networks still rely on left-right consistency based loss, but improved from grey-scale consistency to more advanced feature based consistency, such as structural similarity. However, in real-world datasets, left-right consistency is not fully obeyed by image pairs, e.g., the presence of small object moving, seasonal differences~\cite{us3d}. Song et al.~\cite{song} proposed spatially-adaptive self-similarity to generate robust features, which is dedicated to provide more robust feature representations for stereo matching task in real-world scenarios. DSMNet~\cite{DSMNet} uses a new domain normalisation method and structural filters to improve the generalization ability of unsupervised networks. Wang et al.~\cite{wang} combine the attention mechanism with stereo matching framework and propose PASMNet, a stereo matching network that does not need to define the disparity searching range. 

For stereo matching network specialized for satellite images, less attention has been paid before. Tao et al.~\cite{tao2020edge} proposed edge-sense bidirectional pyramid stereo matching network to solve disparity estimation problem of remote sensing images, and in 2022, they proposed to adopt the confidence-aware unimodal cascade and fusion pyramid network to further improve the stereo matching performance of different scale objects and occluded regions~\cite{tao2022confidence}. Concerned about the difficulties in matching untextured areas and building occlusion in satellite images, He et al.~\cite{he2022hmsm} proposed HMSMNet, which uses termed hierarchical multi-scale structure, assisted by a refine module that fuses the left image with its gradient information to achieve good results on GaoFen-7 satellite data. Wei et al.~\cite{wei2024stereo} introduced attention mechanism to remote sensing stereo matching network to enhance feature representation. In summary, there are very few works focusing on the generalization ability of stereo matching networks, and only a few studies on stereo matching are designed for satellite data. When it comes to unsupervised stereo matching networks tailored for remote sensing, Igeta et al.~\cite{igeta} follow the principle of left-right consistency to construct an unsupervised stereo matching network for VHR satellite images. The unsupervised method does not rely on the disparity ground truth, which reduces the cost of datasets collection to some extent. Currently the accuracy of unsupervised methods is generally worse than supervised methods on various datasets, but there is no work comparing the generalization ability of unsupervised and supervised methods.
\vspace{-7pt}
\subsection{Stereo Matching Dataset}
For both supervised and unsupervised stereo matching networks, deep learning algorithm performance is closely related to quality of training dataset. Currently the generic stereo matching datasets mainly include SceneFlow ~\cite{sceneflow}(synthetic dataset), MiddleBury~\cite{MiddleBury} (structured light generating ground truth), KITTI~\cite{KITTI} (LiDAR generating ground truth). Remote sensing stereo matching datasets, although appeared later, also gradually attracted attention. Patil and Comandur firstly proposed the satellite stereo matching dataset SatSetero with dense disparity ground truth, which collects satellite images from WorldView-3 and WorldView-2~\cite{satstereo}. With increasing demand for timely global-scale 3D mapping, in addition to binocular and trinocular satellite images, incidental satellite images are also used for stereo matching. US3D collected multi-date image pairs from WorldView-3 to construct the dataset~\cite{us3d}. WHU-Stereo uses Chinese GaoFen-7 satellite images from six regions to construct the dataset, which covers a variety of terrain features. Overall, the satellite stereo matching datasets are scarce, the disparity ground truth are collected by LiDAR, which is costly and hard to get across different countries~\cite{whustereo}. The different satellites bring different image resolutions, spectral features, and imaging angles, together with different regional target feature distributions, bring difficulties in generalization. There is both a demand and a difficulty to improve the generalization capability of satellite stereo matching.
\vspace{-7pt}
\subsection{Enhancing Generalization Ability of Stereo Matching Network}
Generalizing from abundant synthetic data trained to real scenario data often fails for stereo matching networks, which brings obstacles for utilizing stereo matching technique in real scenes. Some studies have been conducted to enhance the generalization ability by adaptively adjusting feature learning. Wang et al.~\cite{wang2020improvingdeepstereonetwork} propose to utilize gradient-domain smoothness prior and occlusion reasoning in network to enhance generalization ability to real scenarios after learning synthetic data. Li et al.~\cite{fourier_transform} replace the low-frequency amplitude feature of source domain by that of target domain to adapt network to target style. Some networks have been devoted to learn domain-invariant features~\cite{zhang2020domaininvariant,matchingspace,ITSA}. Cai et al.~\cite{matchingspace} transferred features from RGB color space to the matching space, avoiding over-specialization of specific dataset, enhancing generalization capability. To alleviate the problem of artefacts brought by synthetic data, minimizing sensitivity of latent features to input variations to learn robust and shortcut-invariant features is utilized and improves the test performance on real scenes~\cite{ITSA}. Zhang et al.~\cite{featureconsistency} promoted generalization performance by designing pixel-wise stereo contrastive feature loss to maintain consistency of the matching points features. There are also studies that improve generalization assisted by ancillary information or strengthened information by stereo pairs. Noticing artefacts at boundaries when generalizing to new domain, Pang et al.~\cite{zoomandlearn} employ up-sampled stereo pairs to strengthen details, while using graph Laplacian regularization to smooth out artefacts. ~\cite{guided} adopted sparse but reliable depth information form external source to guide matching network, gaining superior generalization ability. Rao et al.~\cite{rethinking_training} explored the impact of training strategies such as pre-training and data augmentation on the accuracy of model, on small target datasets its impact on generalization performance even exceeds the network structures. 

To conclude, some work has begun to focus on improving the generalization performance of remote sensing stereo matching, and some others have paid attention to key training factors on network performance. However, there has been no work to analyze key factors influencing the generalization performance of satellite stereo matching. Due to the generalization needs and difficulties of satellite data in different domains and the lack of related research, we carry out work on this research topic, to analyze the key factors for the generalization performance of remote sensing stereo matching.

\section{Methods}
\label{sec:methods}
Our work aims to analyze how the feature distributions of different train set and the training manner influence the generalization ability of stereo matching network. Therefore, we selected three representative networks, including CFNet~\cite{CFNet}, HMSMNet~\cite{he2022hmsm}, PASMNet~\cite{wang}, for experiments to verify the key factors hold across all three networks, while also adapting CFNet and PASMNet to satellite images. All network structures used were not drastically altered in this work. Table~\ref{tab:methods} summarizes the evaluated architectures employed and improved for satellite images. Brief descriptions of each network and specialized adaptions to remote sensing images are presented. Then comes the datasets constructed to find the key factors and the construction of the training manners. Finally, we propose the early-stop consistency criterion for unsupervised training manner to help promote generalization performance.
\begin{table*}[!htb]
  \caption{Overview of the selected network architectures.\label{tab:methods}}
  \centering
  \scalebox{1.05}{
    \begin{tabular}{ccccc}
    \hline
     Network & Feature Extraction & Cost Volume & Disparity Calculation & Disparity Refinement    \\
    \hline
     RS-CFNet & Feature Pyramid & \makecell{Feature concatenation\\ and group-wise correlation}	& \makecell{Calculate at each scale while \\offering searching range \\for next scale }& Cascaded structure serves to refine \\
     \hline
     HMSMNet & \makecell{Lightweight Siamese \\Multi-scale Pyramid} & \makecell{Fusion for each adjacent \\cost volume pair}& Calculate at each scale & \makecell{Taking left image, \\its gradients and initial \\disparity to refine}  \\
     \hline
     RS-PASMNet & Hourglass Network & \makecell{Cascaded generation \\following features \\without aggregation}		& \makecell{Based on finest \\parallax-attention map} & \makecell{Hourglass network \\taking left feature and \\initial disparity}  \\
     \hline
 \end{tabular}}
  \end{table*}
\vspace{-7pt}
\subsection{Network Architecture}
\begin{figure*}[!htbp]
  \centering
    \includegraphics[width=7in]{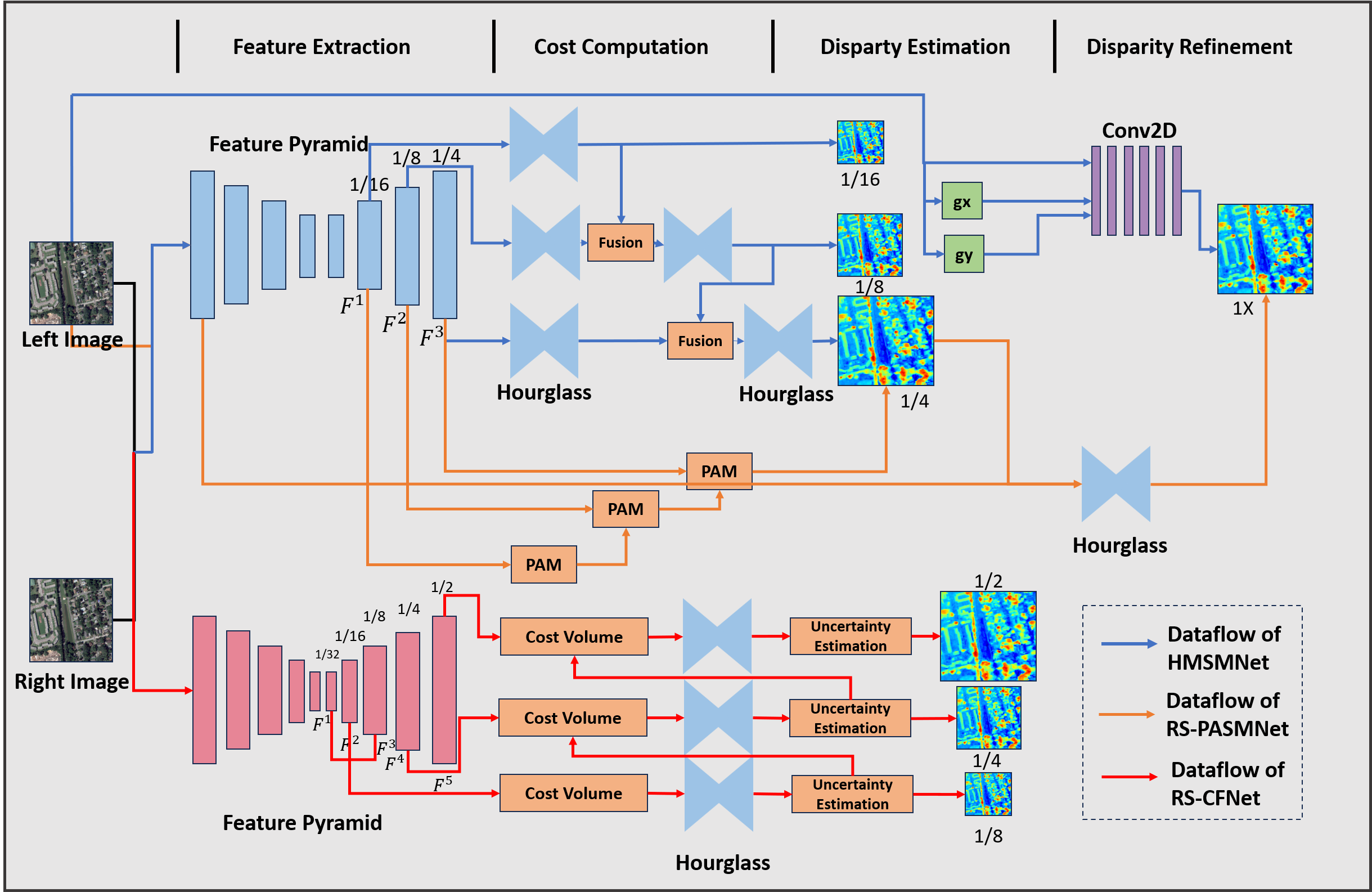}
  \caption{Overview of the selected network architectures. Since all three typical networks follow the basic architecture of classical four modules, they are integrated into the same concept map, using different colored dashes to indicate data flow of different networks: blue for HMSMNet, orange for RS-PASMNet, red for RS-CFNet.}
  \label{fig:RScfnet_structure}
\end{figure*}
Stereo matching network generally follows the four components: feature extraction and fusion, cost computation and aggregation, disparity estimation, disparity refinement. We will brief all the network architectures following this order.
\subsubsection{Feature Extraction and Fusion}
The feature extraction module of stereo matching network is the basement of following matching parts, serves to represent the features of stereo pairs, which will be employed to carry out the following matching cost calculation of disparity candidates. Growing work has noticed the multi-scale features can provide both high-resolution feature representations and high-semantic feature representations, so typically the feature extraction module is supplemented by a feature fusion module. 

Feature extraction and fusion module of CFNet employs a feature pyramid network(FPN) to extract multi-scale features, which shares weight for both left and right image. Features from coarser scales are up-sampled and concatenated with finer-scale features to generate rich representation that captures both global and local image structures. 

For HMSMNet, feature extraction module employs a light-weight Siamese network to extract multi-scale pyramidal features, which also shares weight for both left and right image. Residual blocks and dilation strategy are utilized to gain more robust feature. After feeding multiple features into pooling layers with different pool size, the module finally produces feature maps of three scales: $\mathcal{X}_i \in \mathcal{R}^{H/2^{(i+1)}\times W/2^{(i+1)}\times C}$, where $H\times W$ represents input resolutions and $C$ denotes channel numbers.

Feature extraction for PASMNet is implemented by an hourglass network, then the features of left and right image are fed to PAM module to generate cost volume from coarse to fine. 

\subsubsection{Cost Computation and Aggregation} 
Cost computation after feature extraction and fusion serves to calculate differences between matching pixels. Since pixel-wise cost is easily disturbed by noise and edges, cost aggregation is designed to spread high quality disparity to poorly matched regions. 

For CFNet which utilizes cascaded structure to estimate disparity from coarse to fine, the cost is built correspondingly following feature extraction and fusion at each scale. CFNet uses both feature concatenation and group-wise correlation to generate combination cost volume, which balances multi-scale feature richness and robustness. Cost computation at each scale is given as Eq.~\ref{eq:cfnet-cost}.
\begin{equation}  
	\begin{aligned}
	C_{concat}^{i} (d^i,x,y,f) & = F_{l}^i(x,y)\oplus F_{r}^i(x-d^i,y), \\
	C_{gwc}^{i} (d^i,x,y,f) & = \frac{1}{N_c^i/N_g} \left \langle F_{l}^i(x,y),F_{r}^i(x-d^i,y) \right \rangle, \\
	C_{combine}^i & = C_{concat}^i\oplus C_{gwc}^i  ~,
	\end{aligned}
	\label{eq:cfnet-cost}
\end{equation}
where $\left \langle,\right \rangle$ represents the inner product operation, $\oplus$ denotes the vector concatenation, $F_l^i, F_r^i, N_c$ are the left, right feature maps and the dimension.

To fuse cost volume of multi-resolution, an encoder-decoder-like architecture is used. Encoder-like module consists of four 3D convolution layers to regularize and down-sample cost volume to make alignment, followed by concatenation, feature channel decreasing and down-sampling. Decoder-like module adopts 3D modules to up-sample and refine cost volume. Cost fusion combines information from different scales and enhances generalization ability facing different data domain.

For HMSMNet, cost volume covers both positive and negative disparities. To aggregate and fuse cost volumes, from coarse to fine, cost volume fusion module is adopted for each pair of adjacent cost volumes, coarse-scale cost volume is up-sampled and added to fine-scale cost volume. The channel attention module then takes the summation to generate attention vectors $\alpha$ and $1-\alpha$ to adjust its proportion. The function of cost aggregation is mainly to use large view coarse guidance to rectify the mismatching cost in occluded or untextured regions, maintaining smoothness and continuity of disparity. 

For PASMNet, cost computation and aggregation also follows coarse-to-fine manner, from the first stage to the last, the size of features and matching cost is getting larger. For the first stage, the features of image pairs and matching cost initially set to zero are fed into the first PAM block. After being processed by two convolutions which share the same parameters for left and right feature, the features go through $1\times 1$ convolutions to serve as query feature $\mathcal{Q}$ and key feature $\mathcal{K}$. The matching cost $\mathcal{C}^1_{right\xrightarrow[]{}left}$ is calculated by reshaped $\mathcal{K}$ multiplied with $\mathcal{Q}$. The $\mathcal{C}^1_{left\xrightarrow[]{}right}$ follows the similar way when left and right feature exchanged. Afterwards, the $\mathcal{C}^0_{right\xrightarrow[]{}left}$ and $\mathcal{C}^0_{left\xrightarrow[]{}right}$ are added to $\mathcal{C}^1_{right\xrightarrow[]{}left}$ and $\mathcal{C}^1_{left\xrightarrow[]{}right}$, respectively. $\mathcal{F}^1_{left}$ and $\mathcal{F}^1_{right}$ are added to $\mathcal{F}_{left}$ and $\mathcal{F}_{right}$, respectively. From coarse to fine, thee features and matching cost from preceding PAM block are passed to the succeeding block, similar aforementioned process is conducted to generate finer features and cost. By this process, the cost is aggregated implicitly. 

\subsubsection{Disparity Estimation and Refinement} 
Disparity estimation is used to estimate initial disparity among all the candidates based on former cost computation and aggregation. Considering the unavoidable noises, occlusion, weak or repeated textures in stereo images, the cost after aggregation is not completely accurate. Therefore, the initial disparity obtained by disparity calculation may have some errors, disparity refinement often serves the perspective of improving there errors, using the semantic information in original image pairs or optimisation strategy from coarse to fine to optimise the disparity iteratively on the initial disparity basis. 

For CFNet which has cascaded structure, at each scale, disparity is calculated using softmax operation, given as Eq.~\ref{eq:est_disp}, note that the disparity searching range for the first coarsest scale is defined manually, while for next scales is defined by estimated uncertainty.
\vspace{-7pt}
\begin{equation}
	\hat{d^i} = \sum_{d=-D_{max}^i}^{D_{max}^i} d \times softmax(-c_d^i),
	\label{eq:est_disp}
	\vspace{-10pt}
\end{equation}
where $c_d^i$ represents the initial cost volume, $softmax(-c_d^i)$ denotes the discrete disparity probability distribution. The disparity map is the weighted sum of all disparity indexes.

Uncertainty for each scale is positively correlated with the overall cost, given as standard deviation $\sigma_i$ in Eq.~\ref{eq:uncert_est}. while for pixels with large cost, the uncertainty brought by current disparity estimation is large.
\begin{equation}
\begin{aligned}
    \sigma ^i&=\sqrt{\sum_{d^i}{(}d-\hat{d}^i)^2\times softmax(-c_{d}^{i})}, \\
    \hat{d}^i&=\sum_{d^i}^{}{d}\times softmax(-c_{d}^{i}),
    \end{aligned}
    \label{eq:uncert_est}
\end{equation}
where $c$ represents the predicted cost tensor. Then, the disparity range of the next stage can be computed based on the standard deviation $\sigma_i$ as:
\begin{equation}
\begin{aligned}
d_{max}^{i-1}=\delta (\hat{d}^i+(s^i+1)\sigma ^i+\varepsilon ^i), \\
d_{min}^{i-1}=\delta (\hat{d}^i-(s^i+1)\sigma ^i-\varepsilon ^i),
\end{aligned}
\end{equation}
where $\delta$ denotes the bilinear unsampling operation. $s^i, \varepsilon^i$ are two learnable normalization factors with initial value 0. Finally, based on the uniform spatial sampler, the next-stage disparity can be computed as:
\begin{equation}
\begin{aligned}
d^{i-1} & =d_{min}^{i-1}+n(d_{max}^{i-1}-d_{min}^{i-1})/(N^{i-1}-1),\\
n & \in 0,1,2\dots N^{i-1}-1
\end{aligned}
\end{equation}
where $N^{i-1}$ is internal of disparity search range at stage $i-1$. By adapting the disparity searching range estimation and disparity estimation method to cascaded network, CFNet predicts disparity from coarse to fine, achieving good generalization performance in various domain.

For HMSMNet, disparity estimation is simply computed using sum of disparity candidates' probability, which is negatively correlated with its cost, defined as:$\hat{d^i} = \sum_{d=-D_{max}^i}^{D_{max}^i} d \times softmax(-c_d^i)$. To refine disparity map, the semantic information of original image is used, which means left image and its gradients in $x$ and $y$ direction are fed into refinement module along with original disparity. Gradients is calculated as:
\begin{equation}
	\begin{aligned}
	g_x(x,y) &= I(x+1, y) - I(x-1, y)  \\
	g_y(x,y) &= I(x, y+1) - I(x, y-1) 
	\end{aligned}
	\label{eq:gradients}
\end{equation}
where $g_x$ and $g_y$ denote the gradient along $x$ and $y$ direction, respectively. $I$ represents image intensity. With the finest-scale disparity output as a basis of refinement, the left image and its gradients provide the original semantic information and high-frequency detailed semantics, the refinement module learns the residual to align disparity and the reference image structure. 

For PASMNet, the finest cost is utilized to estimate disparity, firstly fed to softmax layer to generate parallax-attention map $\mathcal{M}\in \mathcal{R}^{\frac{H}{4}\times\frac{W}{4}\times\frac{W}{4}}$ which reveals matching possibility of disparity candidates without maximum limitation. The disparity is estimated as :
\begin{equation}
\hat{D}_{left} = \sum^{W/4-1}_{k=0}k\times \mathcal{M}^3_{right\xrightarrow{}left}(:,:,k)    
\end{equation}
The $\hat{D}_{right}$ follows the similar calculation with $\mathcal{M}^3_{left\xrightarrow{}right}$. The refinement module employs a simple hourglass network to take left feature and initial disparity as input, and produces learnt disparity residual, generating refined disparity with left image structure as guidance.

\subsubsection{Adaption to Satellite Images}
Since CFNet and PASMNet are networks for generalized stereo matching task, the cost volume and disparity estimation constructed set the disparity range to be positive, without taking into account the case of negative disparity, which is, however, a common situation in satellite data. Thus we improve the cost volume module to expand disparity range to cover both positive and negative values. In addition, satellite images and general visual images have large regional target feature distribution differences which are improper to ignore. So in the preprocessing part, we need to calculate mean and variance for each satellite dataset. Both for the training and testing stage in the preprocessing, we use the certain statistical parameters, in order to make the network to achieve stable learning ability. The networks adapted to satellite images are referred to as RS-CFNet and RS-PASMNet in the following sections.
\begin{figure}[!tbp]
    \centering
    \begin{subfigure}{0.4\columnwidth}
        \includegraphics[width=1\columnwidth]{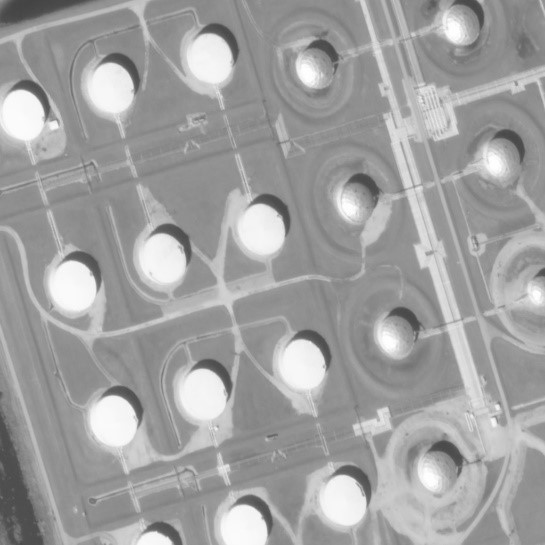}
    \end{subfigure}
    \begin{subfigure}{0.4\columnwidth}
        \includegraphics[width=1\columnwidth]{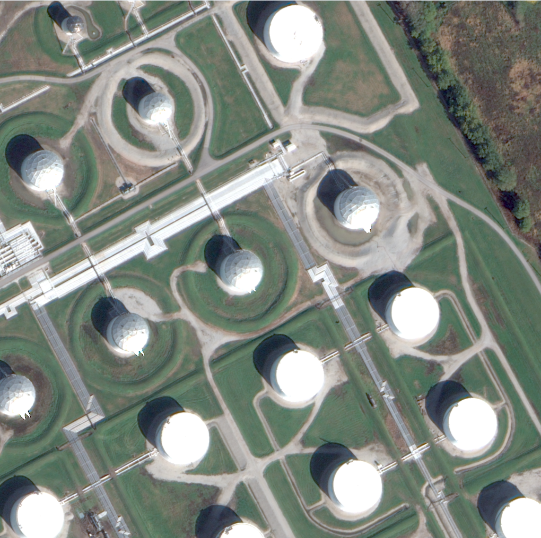}
    \end{subfigure}
    \\
    \begin{subfigure}{0.4\columnwidth}
        \includegraphics[width=1\columnwidth]{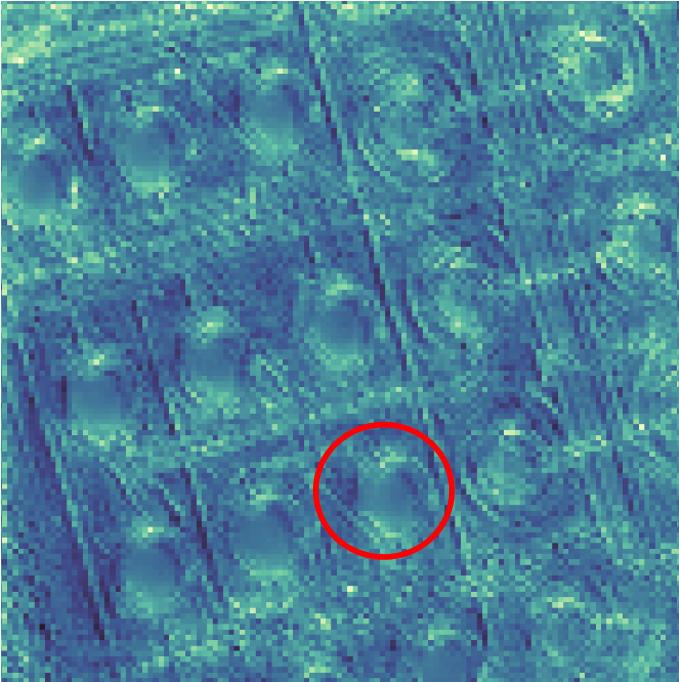}
    \end{subfigure}
    \begin{subfigure}{0.4\columnwidth}
        \includegraphics[width=1\columnwidth]{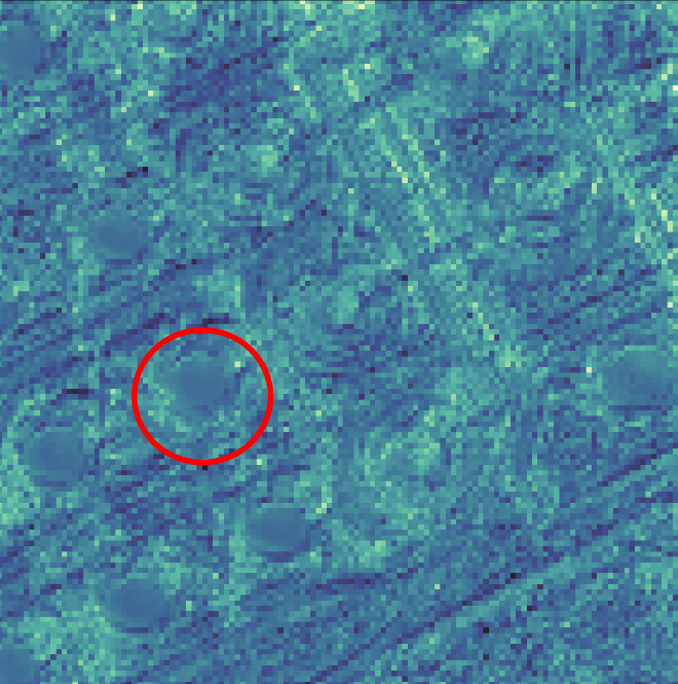}
    \end{subfigure}
  
    \caption{The same target in Omaha imaged by different sensors (Left: GeoEye, Right: WorldView-3, Top:left image, Bottom: corresponding feature visualization of stereo matching network). In the red circles, the features of the same target imaged by different sensors at different angles differ greatly, so the sensor difference also bring possible generalization difficulties.}
    \label{fig:dif_sensor_effect}
\end{figure}

\begin{figure*}[!tbp]
    \centering
    \captionsetup{justification=centering}
    \begin{subfigure}[]{0.3\columnwidth}
        \centering
        \includegraphics[width=1\columnwidth]{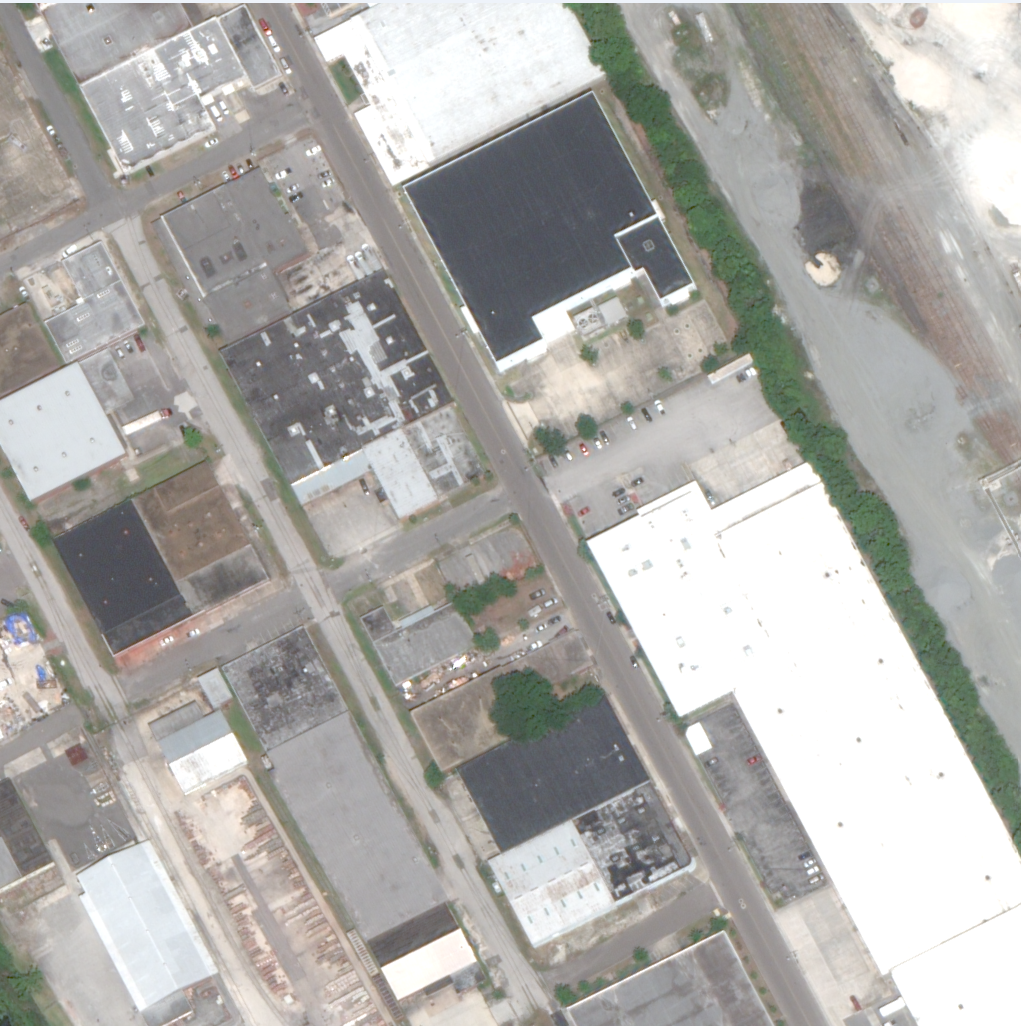}
        \caption{Jacksonville, \\ USA}
    \end{subfigure}
    \begin{subfigure}[]{0.3\columnwidth}
        \includegraphics[width=1\columnwidth]{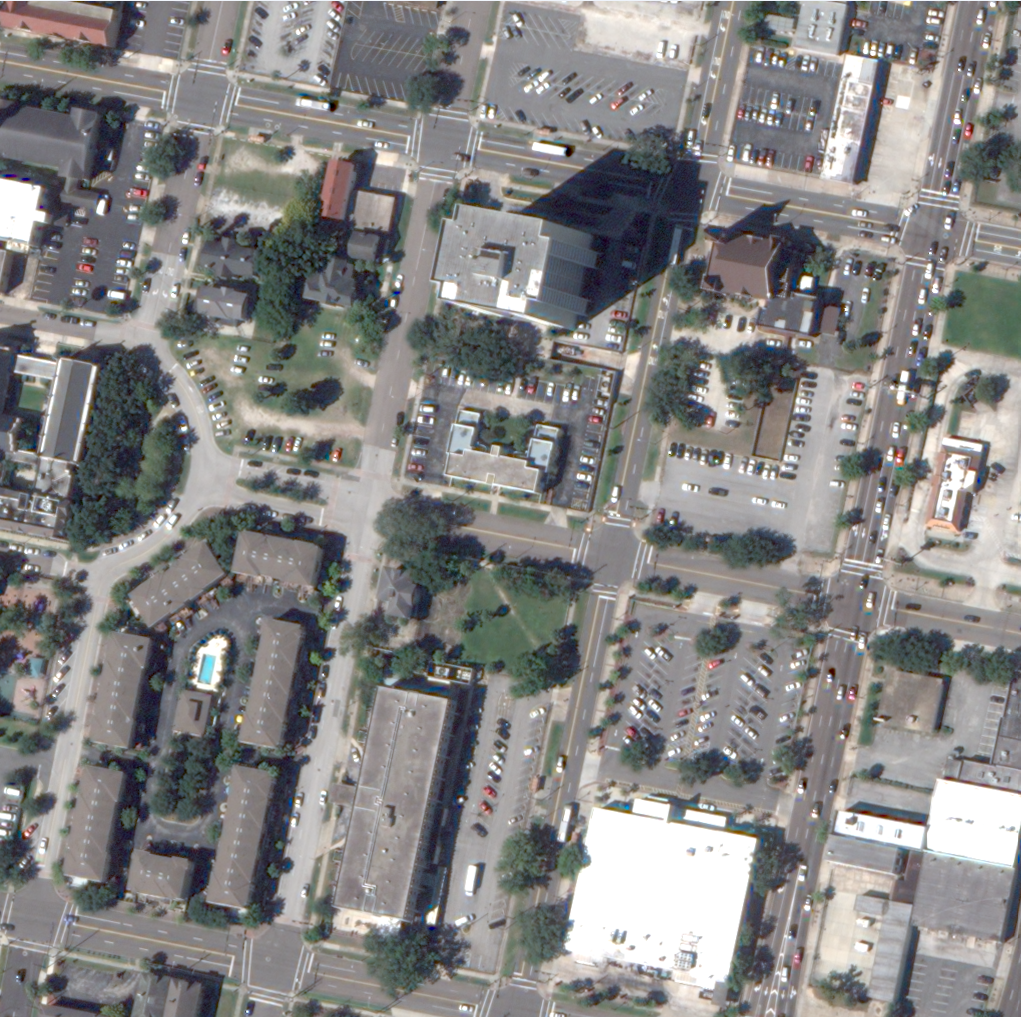}
        \caption{Jacksonville, \\ USA}
    \end{subfigure}
    \begin{subfigure}[]{0.3\columnwidth}
        \includegraphics[width=1\columnwidth]{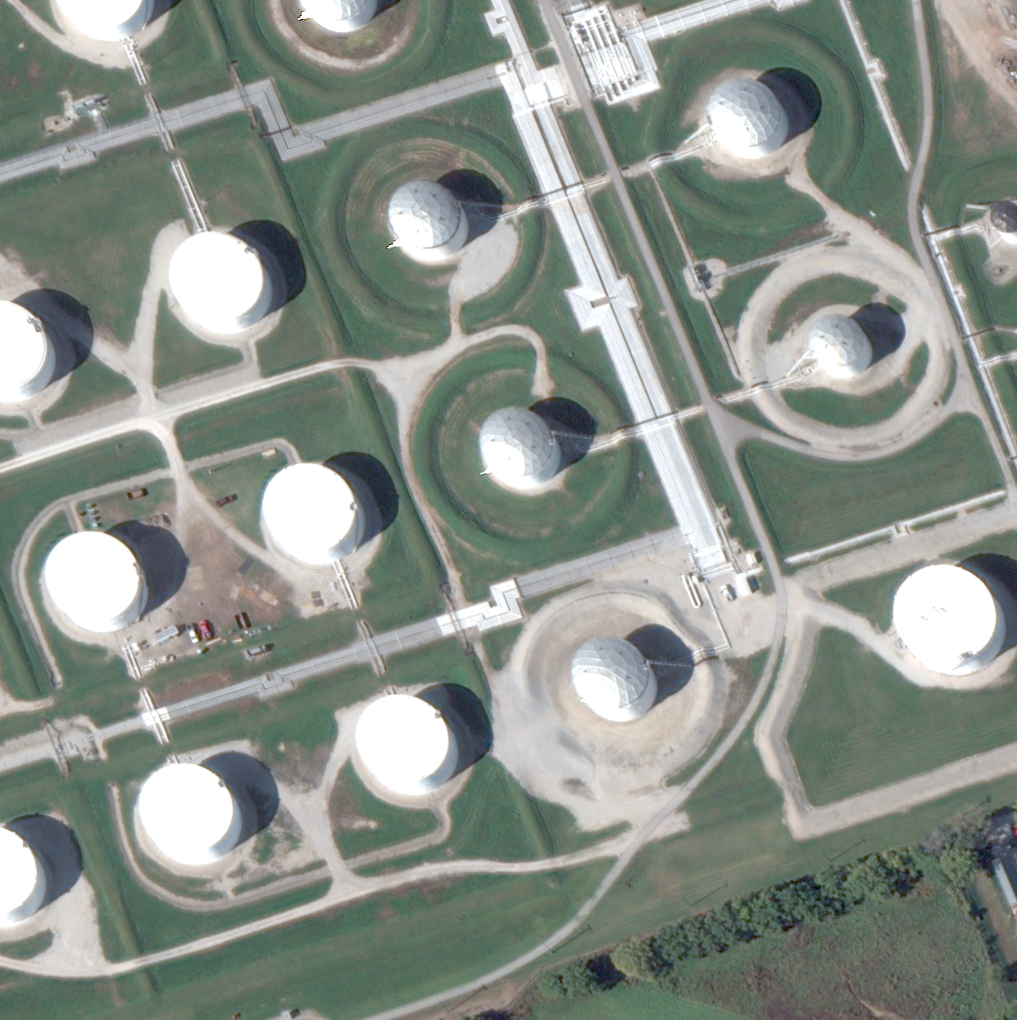}
        \caption{Omaha, \\ USA}
    \end{subfigure}
    \begin{subfigure}[]{0.3\columnwidth}
        \includegraphics[width=1\columnwidth]{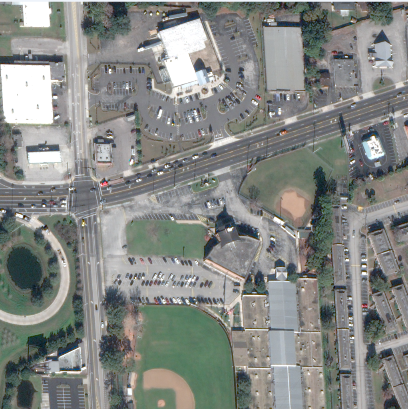}
        \caption{Omaha, \\ USA}
    \end{subfigure}
    \\
    \centering
    \begin{subfigure}[]{0.3\columnwidth}
        \includegraphics[width=1\columnwidth]{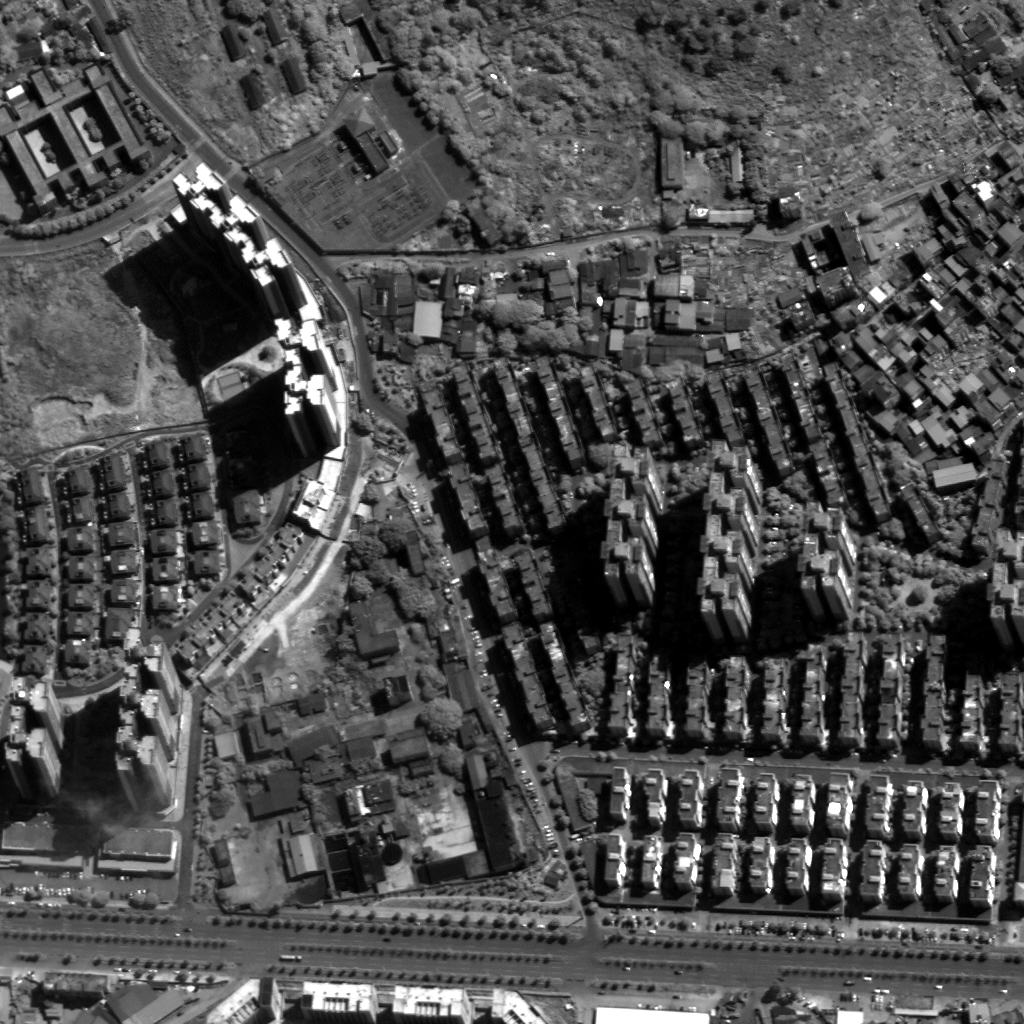}
        \caption{Shaoguan, \\China}
    \end{subfigure}
    \begin{subfigure}[]{0.3\columnwidth}
        \includegraphics[width=1\columnwidth]{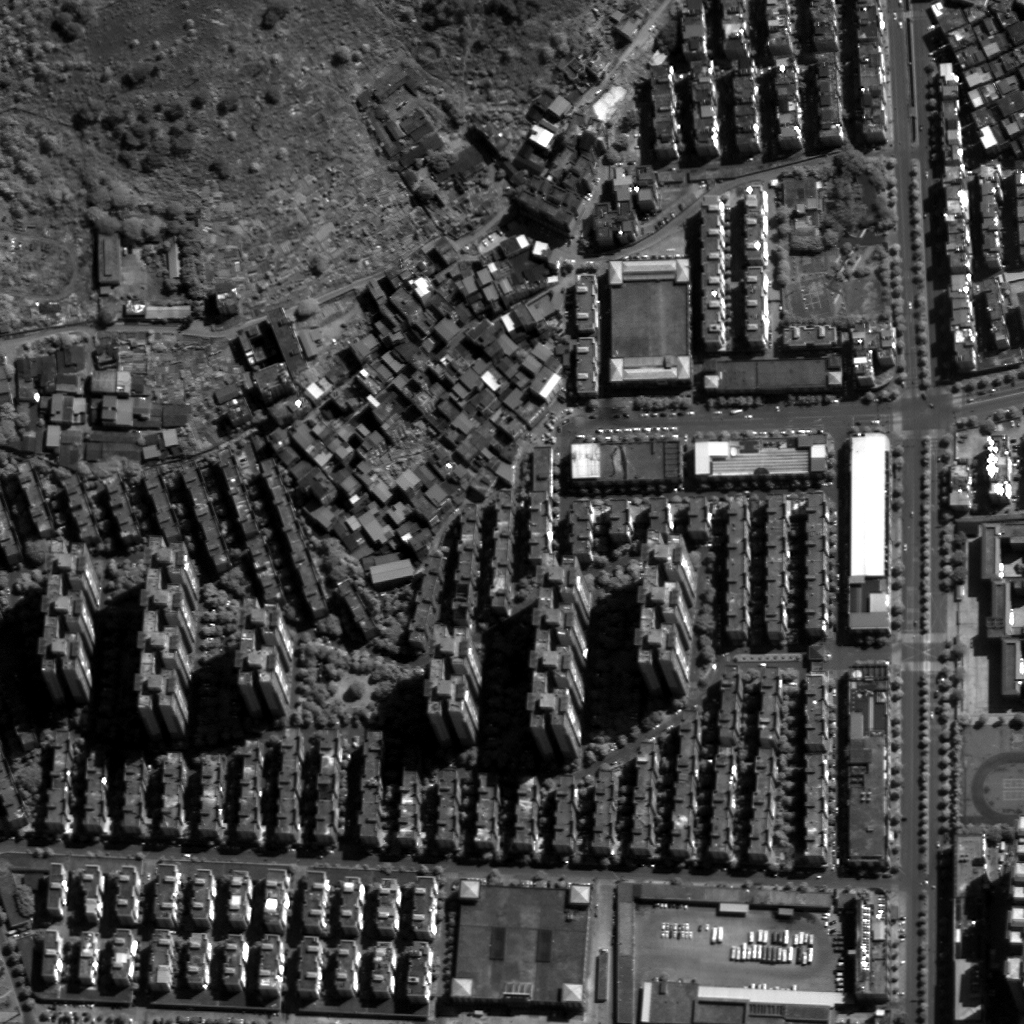}
        \caption{Shaoguan, \\China}
    \end{subfigure}
    \begin{subfigure}[]{0.3\columnwidth}
        \includegraphics[width=1\columnwidth]{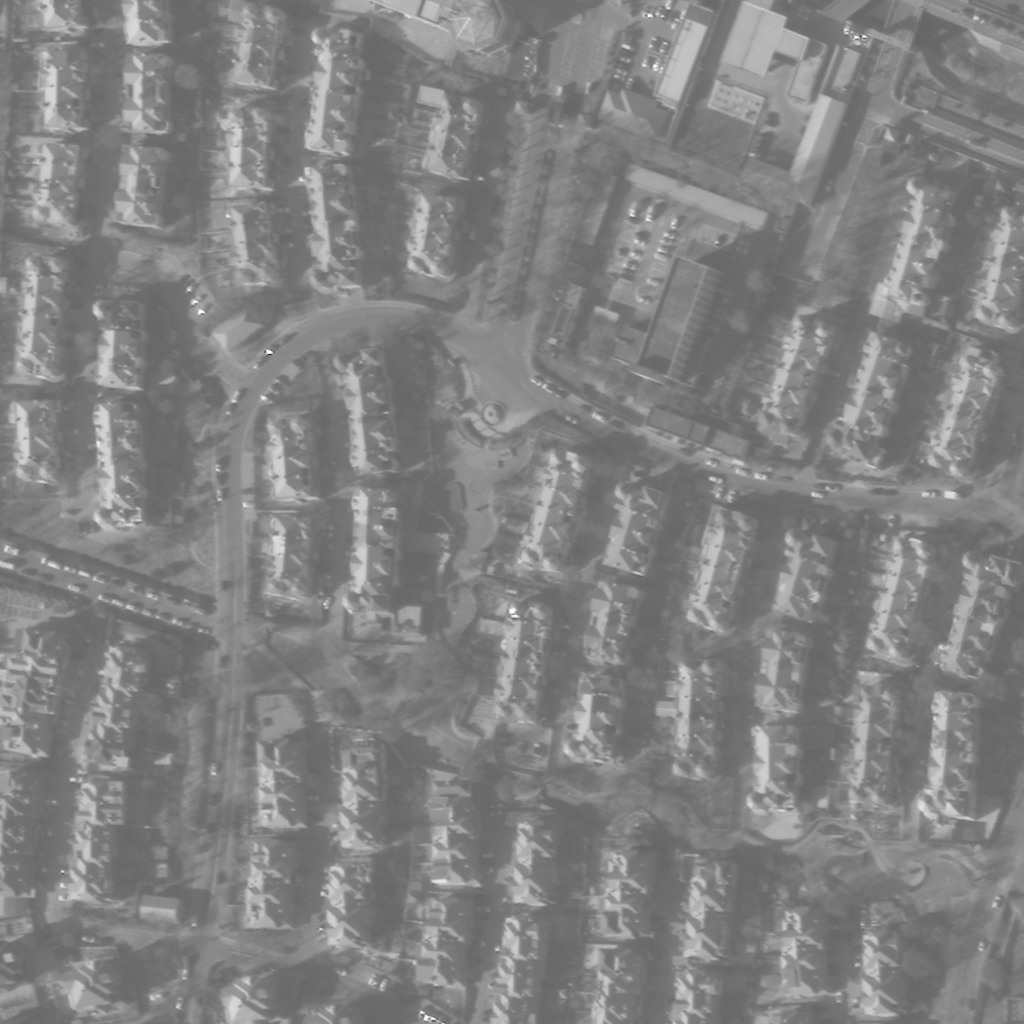}
        \caption{Beijing, \\ \ China}
    \end{subfigure}
    \begin{subfigure}[]{0.3\columnwidth}
        \includegraphics[width=1\columnwidth]{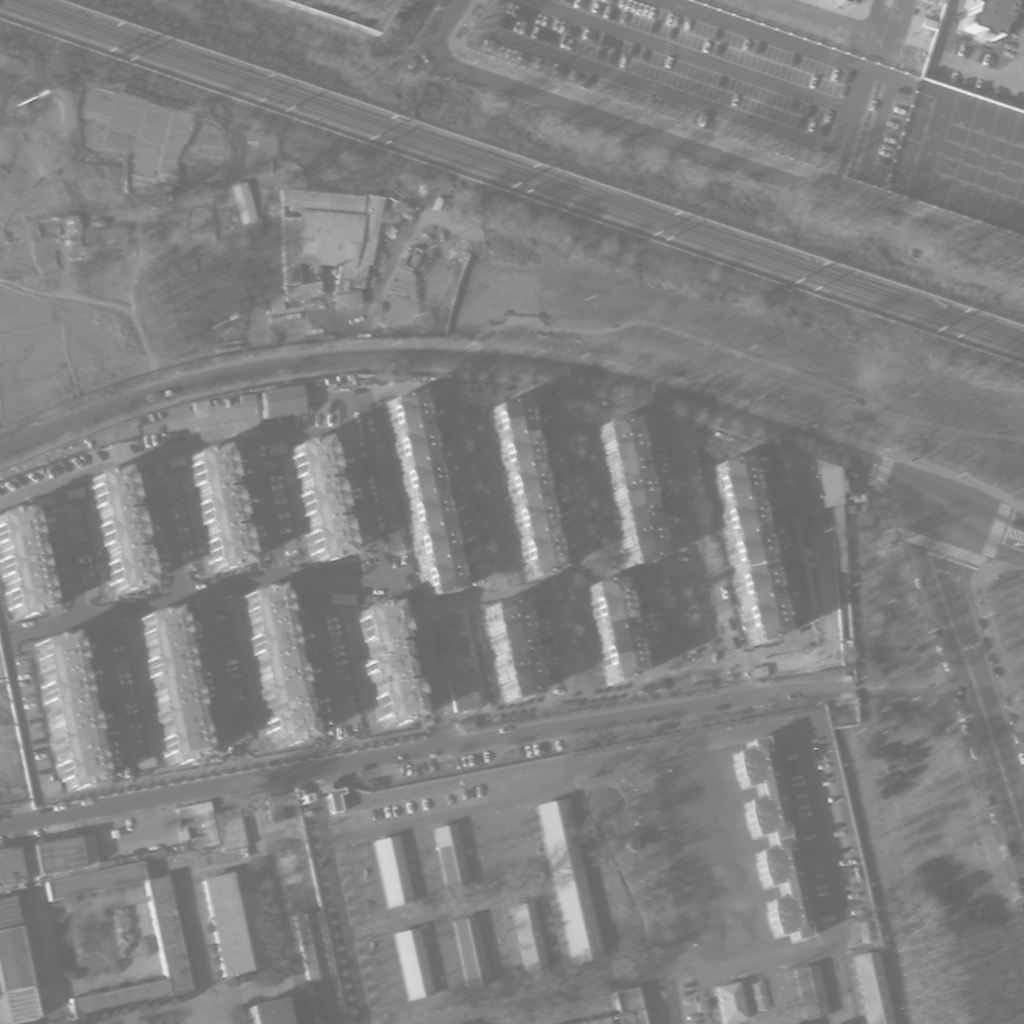}
        \caption{Beijing, \\ \ China}
    \end{subfigure}
    \centering
    \caption{Various regional target features in different cities.}
    \label{fig:dif_region_effect}
\end{figure*}

\begin{table*}[!tbp]
  \caption{Datasets constructed for exploring key factors of generalization.
  *SV-all contains more diversified regional target distribution than SV-bj, while maintaining the same amount of train set image pairs, serving to analyze issue (2).
  \label{tab:dataforsensors}}
  \centering
  \scalebox{1.2}{
  \renewcommand{\arraystretch}{1.2} 
    \begin{tabular}{ccccc}
    \hline
     Purpose&Dataset & Sensor & Region  &Source\\
    \hline
    
     &US3D-train~\cite{us3d} &WorldView-3 & USA:\ Jacksonville	& open-source \\
     train&WHU-Stereo-train~\cite{whustereo} & GaoFen-7 &China:\ Shaoguan\ Kunming\ Yingde\ Qichun	&open-source \\
     
     &SV-bj &SuperView&China:\ Beijing&ours\\ 
     &SV-all* &SuperView&China:\ Beijing\ Jilin\ Zhuhai\ Songshan &ours\\
     \hline
    
     &US3D-test~\cite{us3d} &WorldView-3 & USA:\ Omaha	& open-source \\
     &WHU-Stereo-test~\cite{whustereo} & GaoFen-7 & China:\ Shaoguan\ Kunming\ Yingde\ Qichun\ Wuhan\ Hengyang	&open-source \\
     test&SV-hw &SuperView &USA:\ Hawaii		& ours  \\
     &SV-sd & SuperView & USA:\ San Diego 		&ours  \\
     &GeoEye-omh &GeoEye&USA:\ Omaha&ours\\
     \hline
 \end{tabular}}
  \end{table*}
\subsection{Dataset Constructed for Analysis}
\label{sec:datasetsystem}
Ideally, we construct data with the same regional target feature distribution and from the same sensor as the test set as a training set to reduce generalization difficulty. In practice, however, the construction of stereo image pairs and the acquisition of disparity ground truth are difficult for remote sensing stereo matching tasks, making it not always feasible to construct a dataset for training from the same region or same sensor tested. 

Analyzed from sensors perspectives, we compare the features of the same regional target imaged by different sensors. Given the difference of sensor resolution and sensor channel characteristics, the features extracted by the same stereo matching model appear to be different, as shown in Fig.~\ref{fig:dif_sensor_effect}. That is to say, even testing the same target once appeared in train set, difficulties of generalization may still exist due to different sensors. Analyzed from regional target feature distribution, the features of different cities vary significantly, while transnational ground features vary even severer, which also brings difficulty for generalization, as shown in Fig.~\ref{fig:dif_region_effect}.

Based on our definition of domain, which indicates the same domain means from the same city and the same sensor, we face the promotion needs of cross-domain generalization performance. Here, we design to specify three issues that may affect generalization capability from train set perspective. 

(1)~Is using data from the same sensor for training helpful for generalization? 

(2)~Is using data with diversified regional target distribution helpful for generalization? 

(3)~Is using data with similar regional target distribution helpful for generalization?
  
To study the above questions, we build the following datasets as Table.~\ref{tab:dataforsensors} to conduct experiments. We train separately for each training set while results are tested in all test sets. We designed data containing abundant various regional target feature distribution, as well as data from different sensors, and prepared the corresponding test sets as well, whereby we can analyze the impact of sensors and the target feature distribution on the generalization performance. To address the issue of whether the diversity of regional target distribution can help improve generalization ability, we designed SV-all, compared to SV-bj, which contains exact the same amount of training samples, while covering richer set of cities with different architectural styles, serving to analyze this issue.

\subsection{Training Manners}
\label{sec:cfnet_loss}
Unsupervised learning methods avoid the need of ground truth collection when constructing data set, while supervised learning methods utilize ground truth as supervision information, thus achieving superior performance on the same test domain compared to unsupervised methods. However, when testing on new domain, it is unclear about the generalization ability comparisons between supervised training manner and unsupervised training manner. Here, we design an experimental procedure to analyze the effect of training manner on generalization performance.

Since we assess both supervised and unsupervised learning approaches, we present the supervised and unsupervised losses separately.

For supervised loss built on the basis of disparity ground truth, smooth L1 loss for each scale is utilized, defined as Eq.~\ref{eq:sup}.  
\vspace{-7pt}
\begin{equation}
\begin{aligned}
\mathcal{L}_{sup}&=\frac{1}{N} \sum_{i,j}^{}\omega_j smoothL1(\hat{d}^{i,j},{d}^{i,j})\\
\label{eq:sup}
\end{aligned}
\vspace{-7pt}
\end{equation}
For unsupervised loss, the loss is constructed following left-right consistency criterion. The loss function is the same for each scale, denoted $L_S$. The full loss is the sum of individual scale losses, i.e., $\mathcal{L}_{unsup}=\sum^{}_{s}\mathcal{L}_s$.
The $\mathcal{L}_{s}$ combines three terms, given as:
\begin{equation}
\mathcal{L}_s=\lambda_{p}\mathcal{L}_{ap}+\lambda_{census}\mathcal{L}_{census}+\lambda_{sm}\mathcal{L}_{sm}
\end{equation}
\begin{small}
\begin{equation}
	\begin{aligned}
	\mathcal{L}_{p} & = \frac{1}{N} \sum_{i,j}^{}\alpha \frac{1-SSIM(I_{i,j}^{l} \odot (1 - O_{i,j}^{l}),\tilde{I}_{i,j}^{l} \odot (1 - O_{i,j}^{l}) ) }{2}  \\
	&+ (1-\alpha )\left \| I_{i,j}^{l} \odot (1 - O_{i,j}^{l}),\tilde{I}_{i,j}^{l} \odot (1 - O_{i,j}^{l})  \right  \|   
	\end{aligned}
	\label{eq:re}
\end{equation}
\end{small}

$L_{p}$ promotes consistency between reference image and its reconstructed image in non-occluded areas as Eq.~\ref{eq:re}, while we use a simple formulation to detect occluded areas by distance of forward-backward disparities, defined as Eq.~\ref{eq:occ}:
\begin{equation}
|d_i^f(j)+d_i^b(d^f_i(j))|^2<\tau_i 
    \label{eq:occ}
\end{equation}
where $d^f(j)$ represents the forward disparity based on left image,  $d^b(j)$  represents backward disparity  based on right image,  $O_{i,j}^{l}$ is the predicted occlusion map. $\odot$ denotes element-wise multiplication, $\alpha$ is set to 0.85. $\tau_i=[5,2,1]$ according to the general pattern of satellite images.

$\mathcal{L}_{census}$ is also adopted to reduce reconstruction errors while it has better robustness to various brightness and shadow disturbances in satellite images from diverse datasets. We utilize Census transform($\tau$) with a patch size of 7 pixels and Charbonnier penalty function($\rho$) to implement $L_{census}$, Hamming Distance denoted as $\Gamma$. The formulation is as follow:
\begin{equation}
	\mathcal{L}_{census}  = \frac{1}{N} \sum_{i,j}^{} \rho[\Gamma(\tau(I_{i,j}^{l}) ,\tau(\tilde{I}_{i,j}^{l}))] \odot (1 - O_{i,j}^{l}) ) 
\end{equation}

$\mathcal{L}_{sm}$ is utilized to smooth disparity with $L_2$ penalty on disparity gradients $\partial{d}$ and image gradients $\partial{I}$, denoted as:
\begin{equation}
    \mathcal{L}_{sm} = |\partial{_x}d_L|e^{-|\partial{_x}I_L|}+|\partial{_y}d_L|e^{-|\partial{_y}I_L|} 
\end{equation}

For network based on attention mechanism like PASMNet, loss denoted as $\mathcal{L}_{unsup\_attention}$ only serves for unsupervised learning approach, it combines three parts including photometric loss $\mathcal{L}_p$, smooth loss $\mathcal{L}_{sm}$ and unsupervised loss $\mathcal{L}_{PAM}$. As $\mathcal{L}_p$ and $\mathcal{L}_{sm}$ have been discussed in section~\ref{sec:cfnet_loss}, we only detail $\mathcal{L}_{PAM}$ here. The $\mathcal{L}_{PAM}$ for each scale is defined as:
\begin{equation}
    \mathcal{L}^s_{PAM} = \mathcal{L}^s_{PAM-p}+ \lambda_{PAM-s}\mathcal{L}^s_{PAM-s} +\lambda_{PAM-c}\mathcal{L}^s_{PAM-c}
\end{equation}
$\mathcal{L}^s_{PAM-p}$ is built on the idea to make reference image and its reconstruction being consistent, similar to photometric loss, but it is based on parallax-attention map, defined as:
\begin{equation}
    \begin{aligned}
	\mathcal{L}^s_{PAM-p}  = \frac{1}{N^s_{left}} &\sum_{i,j}\left\| I_{i,j}^{l} , (M^s_{right \xrightarrow{}left}\otimes {I}_{i,j}^{r} )\right\| \\
	&\odot (1 - O_{i,j}^{l})   
	\end{aligned}
\end{equation}
where $N^s_{left}$ are the numbers of non-occluded pixels in left image, and $O^{l}(i,j)$ is occluded pixels inferred by $M_{left\xrightarrow{}right}$, defined as Eq.~\ref{eq:pamocclu}. For right image, similar formulation is applied, while left and right images exchanged.
\begin{equation}
O^{l}(i,j) = \left\{
\begin{aligned}
        	 1,&\qquad  if \sum_{k\in[1,W]}M_{left\xrightarrow{}right}(i,k,j) \leq \tau \\
        	 0,&\qquad  otherwise \\
\end{aligned} 
\right.
\label{eq:pamocclu}
\end{equation}
$\mathcal{L}^s_{PAM-s}$ serves to regularize parallax-attention map, defined as:
\begin{equation}
\begin{aligned}
    \mathcal{L}^s_{PAM-s} =&\frac{1}{N^s}\sum_{M^s}\sum_{i,j,k}(\left\| M^s(i,j,k) - M^s(i+1, j,k)\right\|) \\
    &+\left\| M^s(i,j,k)- M^s(i,j+1,k+1)\right\|
\end{aligned}
\end{equation}
where $M^s\in\{M^s_{left\xrightarrow{}right},M^s_{right\xrightarrow{}left}\}$, the terms are used to make attention map consistent in vertical and horizontal direction, respectively.
$\mathcal{L}^s_{PAM-c}$ is a cycle consistency loss introduced to keep left and right attention map consistent. Following the idea that $M^s_{left\xrightarrow{}right\xrightarrow{}left}$ and $M^s_{right\xrightarrow{}left\xrightarrow{}right}$ should be as identity matrices ideally, the cycle loss is defined as:
\begin{equation}
    \begin{aligned}
        \mathcal{L}^s_{PAM-c} = \frac{1}{N^s_{left}} &\sum_{i,j} \left\| [M^s_{left\xrightarrow{}right\xrightarrow{}left}(i,j)-I^s_{i,j}]\right \| \\
        &\odot (1 - O_{i,j}^{l}) 
    \end{aligned}
\end{equation}
where $I^s_{i,j}$ is identity matrix, note that for right image, the cycle loss is similar. Overall, the whole unsupervised loss for PASMNet is defined as Eq.~\ref{eq:pamunsupfull}, where $\omega_s=[0.2,0.3,0.5]$ to balance optimization of each scale.
\begin{equation}
    \begin{aligned}
        \mathcal{L}_{unsup} = \mathcal{L}_p +\lambda_s\mathcal{L}_s+\lambda_{PAM}\omega_s\mathcal{L}^s_{PAM}
    \end{aligned}
    \label{eq:pamunsupfull}
\end{equation}

Since RS-CFNet and HMSMNet are employed for both supervised and unsupervised training manners, they use $\mathcal{L}_{sup}$ and $\mathcal{L}_{unsup}$ for supervised and unsupervised training separately, note that the weight to balance optimizing of each scale is sequentially $\omega=[0.5, 1.0, 2.0]$ and $\omega=[0.5, 0.7, 1.0 , 0.6]$ from coarse to fine for RS-CFNet and HMSMNet. For RS-PASMNet which depends on attention mechanism, it utilizes attention-based unsupervised loss $\mathcal{L}_{unsup\_attention}$.
\vspace{-7pt}
\subsection{Training Setting}
\subsubsection{Basic Experiment Setting}
On the basis of the network structure presented, we describe the specific experiment settings here. For a fair comparison, for all models, our training process is performed on a single NVIDIA Tesla V100, and all testing is performed on a NVIDIA Titan-RTX.

For RS-CFNet, we trained 20 epochs from scratch using the Adam optimiser ($\beta_1$ = 0.9, $\beta_2$ = 0.999) on SceneFlow dataset with learning rate of 0.001 and batch size as 8, which served as the pre-trained model. On this basis, we continue training with formal trainset cropped to $512\times512$ pixels. When we train in a supervised manner, we train with a learning rate of 0.0001 until the validation set loss converges, while for unsupervised manner, we use a learning rate of 1e-6 until the consistency criterion stopping condition is met. For models unsupervised training in ``w/o" mode which means without pre-trained model, we train from scratch with an initial learning rate of 1e-4, dropping by half every 5 epochs until 1e-7 holds.

For HMSMNet, we trained from scratch using the Adam optimiser ($\beta_1$ = 0.9, $\beta_2$ = 0.999). For both supervised and unsupervised training methods, our learning rate was 0.001 and decreased by half every 10 epochs.

For RS-PASMNet, we trained from scratch using the Adam optimiser ($\beta_1$ = 0.9, $\beta_2$ = 0.999) with The learning rate 0.001 and for every ten epochs the learning rate decreased by one-tenth until 1e-7 is maintained. The training is done until the loss converges.

For initial disparity searching range, the full global range is covered by RS-PASMNet so there is no need to set the searching range. For HMSMNet, we set it to [-96, 96] for the US3D dataset and [-128, 64] for other datasets. For RS-CFNet, we set [-128, 128] for all datasets.

Regarding the metrics of test, Average Endpoint Error (EPE) and the Fraction of Erroneous Pixels (D1) are selected. EPE is the average of Euclidean distance between estimated disparity and ground truth disparity. D1 is the percentage of pixels having absolute disparity error of more than 3 pixels while also larger than 5\% of the ground truth disparities. 
\subsubsection{Early-stop Consistency Criterion}
\label{earlystop}
Given the scarcity of remote sensing stereo matching datasets and the huge cost of building them, it would be helpful to be able to utilise a large number of synthetic stereo matching datasets as pre-training data. However, it has been shown in abundant experiments that when unsupervised training is performed based on a pre-trained model, the results may appear to improve before deteriorating. We propose the consistency criterion to stop the unsupervised training at the proper time and keep the best model.

\begin{figure}[!t]
  \centering
    \includegraphics[width=3.5in]{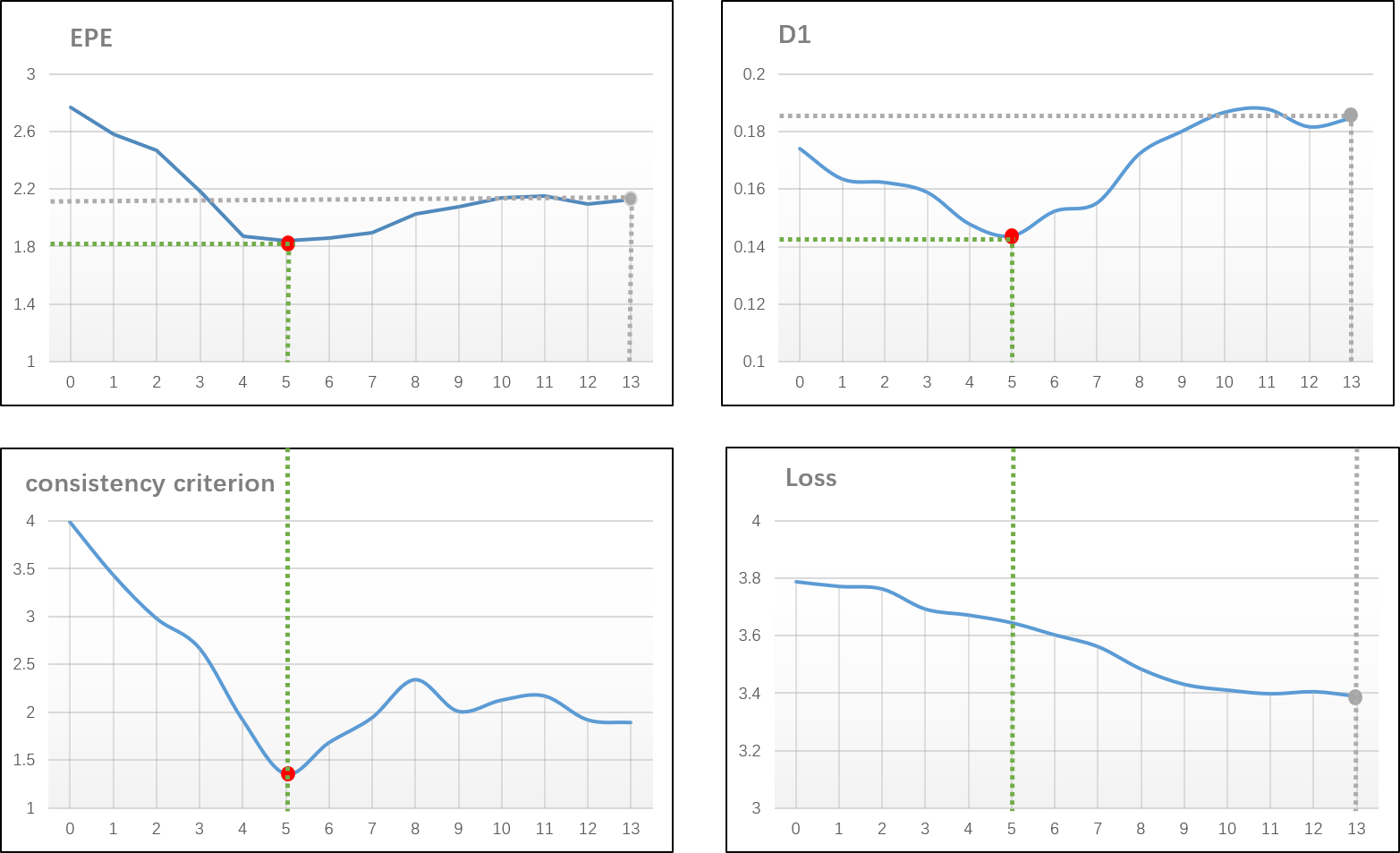}
  \caption{The curves of EPE, D1, consistency criterion and Loss when training on US3D dataset with unsupervised training manner based on pre-trained model, indicating the consistency criterion is effective for properly stopping the training.}
  \label{fig:igarssCE}
\end{figure}

\begin{figure}[!t]
  \centering
    \includegraphics[width=3.5in]{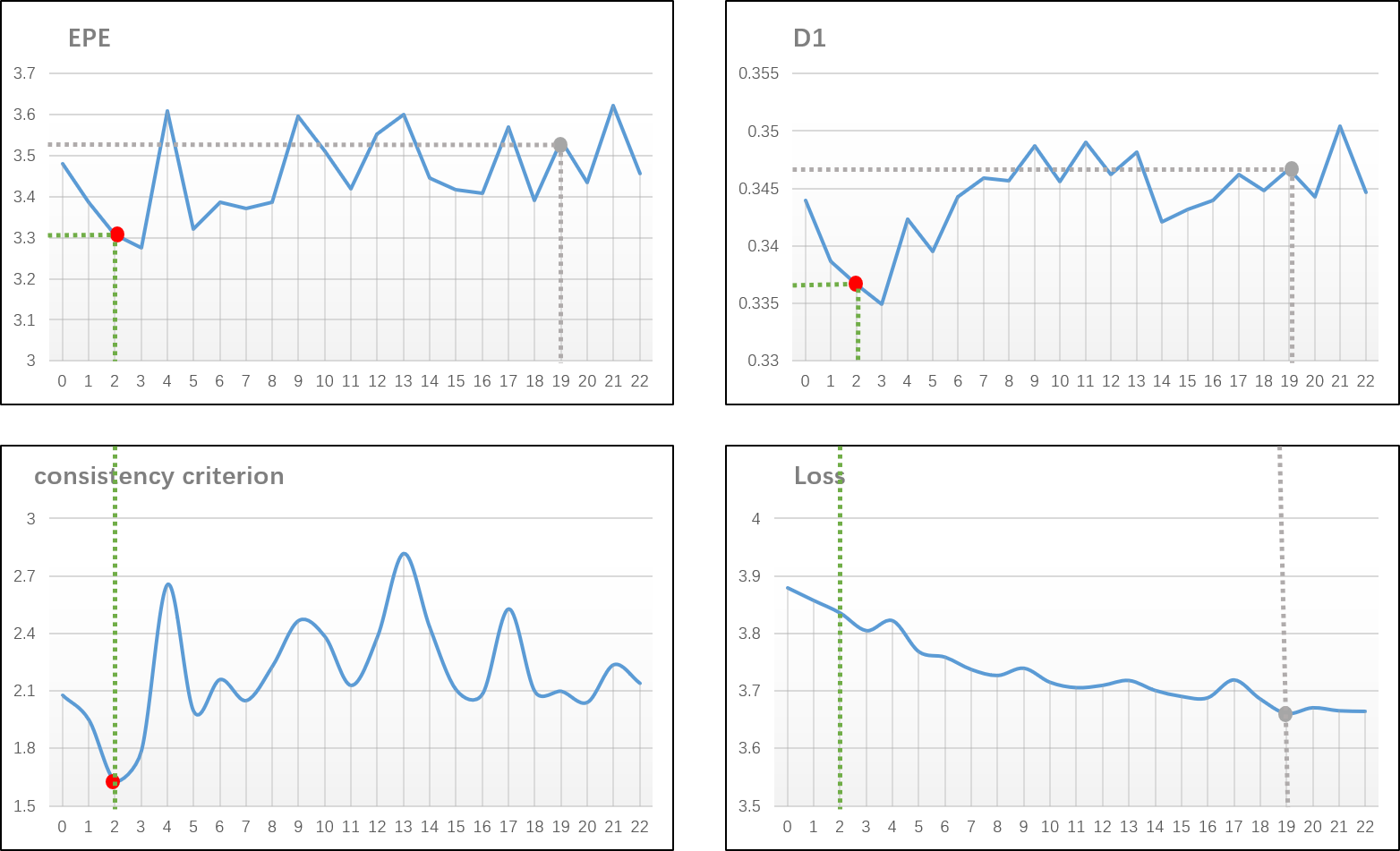}
  \caption{The curves of EPE, D1, consistency criterion and Loss when training on WHU dataset with unsupervised training manner based on pre-trained model, indicating the consistency criterion is effective for properly stopping the training.}
  \label{fig:whuCE}
\end{figure}

Taken advantage of the good left-right consistency of synthetic datasets, the model pre-trained on the synthetic dataset SceneFlow can be used as the pre-trained model, and fine-tuning on the satellite dataset on this basis can achieve better results. The left-right consistency criterion is defined as Eq.~\ref{eq:lrconsis}, revealing the extent of inconsistency of left and right disparity in model inference.
\begin{equation}
\begin{aligned}
    CE =& \frac{1}{N^l_{i,j}} &\sum_{i,j}\left| \hat{d}_{i,j}^{l} + \mathcal{W}(\hat{d}_{i,j}^{r}, \hat{d}_{i,j}^{l}))\right| \\
    +&\frac{1}{N^r_{i,j}} &\sum_{i,j}\left| \hat{d}_{i,j}^{r} + \mathcal{W}(\hat{d}_{i,j}^{l}, \hat{d}_{i,j}^{r}))\right|
\end{aligned}
\label{eq:lrconsis}
\end{equation}
$\hat{d}^{l}$ and $\hat{d}^{r}$ are predicted left and right disparity, $\mathcal{W}$ is warping operation. The fine-tuning model stops training as soon as the CE criterion rises. Trends in consistency criterion, unsupervised loss, EPE and D1 are shown in Fig.~\ref{fig:igarssCE}, Fig.~\ref{fig:whuCE}, the horizontal axis represents the training epoch. When training to loss convergence, the best model will miss the actually best epoch, while the consistency criterion constrains the model to stop at the most consistent epoch. There is a strong correlation between the real performance and consistency on data sets like US3D, WHU-Stereo datasets. Analyzed as follows, although the satellite image stereo matching dataset used epipolar calibration, the differences in the left and right images due to shadows, radiometric differences, small object motion, etc. still exist. So that the left-right consistency is not satisfied at every pixel, thus may pose a problem for unsupervised methods relying on left-right consistency. Fortunately, the synthetic dataset strictly follows the left-right consistency to generate, and the unsupervised learning using it as the pre-training basis can achieve the optimal results by subsequently taking the epoch with the best left-right consistency.

\section{Experimental Results}
\label{sec:exp}
In order to study the effect of training datasets and training methods on the generalization performance, we conducted experiments on the selected three network structures to exclude the interference of network structures, prepared four training sets and five test sets, using both supervised and unsupervised training methods. The specific experimental analyzes are described in detail below.

\begin{table*}[!htbp]
  \caption{Results about the influence of training dataset on generalization performance. To compare the generalization performance between different trainset in the setting of the same defined network structure and training manner, we label the best results as pink and the second best results as light blue, while all results rested in the same domain as the train set are not concerned in comparison. The model with best generalization performance on multiple test set is marked as pink. \label{tab:modelperformance}}
  \centering
  \scalebox{1.05}{
\renewcommand{\arraystretch}{1.0} 
    \begin{tabular}{ccccccccccccc}
    \toprule
     \multirow{2}{*}{Model} &\multirow{2}{*}{Manner} &\multirow{2}{*}{Trainset} & \multicolumn{2}{c}{US3D-test} & \multicolumn{2}{c}{WHU-Stereo-test} & \multicolumn{2}{c}{SV-hw} & \multicolumn{2}{c}{SV-sd} &\multicolumn{2}{c}{GeoEye-omh} \\
     \cmidrule(){4-13}
    &&& EPE & D1 & EPE & D1& EPE & D1 & EPE & D1 &EPE & D1\\
    \midrule
    RS-CFNet&supervised&\cellcolor{mypink}US3D-train&1.47&8.23&3.48&33.53&\cellcolor{mypink}2.23&\cellcolor{mypink}20.97&\cellcolor{mypink}2.67&\cellcolor{mypink}22.67&\cellcolor{mypink}1.97&\cellcolor{mypink}15.73\\
    RS-CFNet&supervised&WHU-train&1.90&15.24&2.54&25.69&2.81&24.18&3.06&25.77&2.58&18.88\\
    \hline
    RS-CFNet&unsupervised&\cellcolor{mypink}US3D-train&1.96&14.51&3.26&32.31&\cellcolor{mypink}2.45&\cellcolor{mypink}21.83&\cellcolor{mypink}2.61&\cellcolor{mypink}23.24&\cellcolor{myblue}2.36&\cellcolor{mypink}17.02\\
    RS-CFNet&unsupervised&WHU-train&\cellcolor{myblue}2.19&\cellcolor{myblue}19.75&3.32&33.87&2.62&22.46&\cellcolor{myblue}2.81&25.04&2.40&17.73\\
    RS-CFNet&unsupervised&SV-bj&\cellcolor{mypink}2.18&\cellcolor{mypink}17.32&\cellcolor{mypink}3.13&\cellcolor{mypink}32.04&2.65&\cellcolor{myblue}22.04&\cellcolor{myblue}2.81&\cellcolor{myblue}24.76&\cellcolor{mypink}2.34&\cellcolor{myblue}17.23\\
    RS-CFNet&unsupervised&SV-all&2.97&20.94&\cellcolor{myblue}3.24&\cellcolor{myblue}32.24&\cellcolor{myblue}2.53&22.09&\cellcolor{myblue}2.81&25.13&2.50&17.75\\
    \hline
     RS-CFNet&unsupervised-w/o\ pre.&US3D-train&2.24&20.94&6.46&40.94&\cellcolor{mypink}2.54&\cellcolor{mypink}23.94&\cellcolor{mypink}2.97&\cellcolor{mypink}27.18&\cellcolor{mypink}2.18&\cellcolor{mypink}19.19\\
    RS-CFNet&unsupervised-w/o\ pre.&WHU-train&3.25&30.12&3.59&36.70&4.09&30.00&3.26&30.30&6.06&29.53\\
    
    \hline
    HMSMNet&supervised&\cellcolor{mypink}US3D-train&1.78&12.74&6.07&43.78&\cellcolor{mypink}2.76&\cellcolor{mypink}25.56&3.14&\cellcolor{mypink}27.22&\cellcolor{mypink}2.25&\cellcolor{mypink}18.42\\
    HMSMNet&supervised&WHU-train&2.21&19.98&2.89&29.13&3.21&29.67&\cellcolor{mypink}3.12&28.07&2.74&23.34\\
    \hline
    HMSMNet&unsupervised&\cellcolor{mypink}US3D-train&2.09&18.34&5.50&38.63&\cellcolor{mypink}2.40&\cellcolor{mypink}23.29&\cellcolor{mypink}2.84&\cellcolor{mypink}25.72&\cellcolor{mypink}2.04&\cellcolor{mypink}17.73\\
    HMSMNet&unsupervised&WHU-train&2.37&23.13&3.87&36.73&2.50&24.16&2.95&26.66&2.31&20.16\\
    \hline
    RS-PASMNet&unsupervised&\cellcolor{mypink}US3D-train&2.47&24.81&8.94&61.27&\cellcolor{mypink}3.75&\cellcolor{mypink}37.08&\cellcolor{mypink}4.13&\cellcolor{mypink}44.52&\cellcolor{mypink}2.94&\cellcolor{mypink}28.95\\
    RS-PASMNet&unsupervised&WHU-train&4.20&52.41&7.17&55.70&5.43&57.30&5.04&53.24&5.39&58.81\\
     \bottomrule
 \end{tabular}}
 \label{tab:performance}
  \end{table*}

\subsection{For training dataset}

\begin{table}[!htbp]
  \caption{Model setting.\label{tab:modelsforsimilar}}
  \centering
  \scalebox{1.3}{
\renewcommand{\arraystretch}{1.2} 
    \begin{tabular}{ccc}
    \hline
     model & trainset & manner \\
    \hline
    RS-CFNet&US3D-train&supervised\\
    RS-CFNet&WHU-Stereo-train&supervised\\
    RS-CFNet&US3D-train&unsupervised\\
    RS-CFNet&WHU-Stereo-train&unsupervised\\
    RS-CFNet&SV-bj&unsupervised\\
    RS-CFNet&SV-all&unsupervised\\
    \hline
    HMSMNet&US3D-train&supervised\\
    HMSMNet&WHU-Stereo-train&supervised\\
    HMSMNet&US3D-train&unsupervised\\
    HMSMNet&WHU-Stereo-train&unsupervised\\
    \hline
    RS-PASMNet&US3D-train&unsupervised\\
    RS-PASMNet&WHU-Stereo-train&unsupervised\\
     \hline
 \end{tabular}}
  \end{table}
For all three networks including RS-CFNet, HMSMNet, RS-PASMNet, for both supervised on unsupervised training manner, we conduct thorough experiments on datasets listed, to study effects of training dataset on generalization performance, while excluding the influence of network structure and training manner. All the settings of trained models are listed in Table.~\ref{tab:modelsforsimilar}.

All the performance of evaluated models is shown in Tab.~\ref{tab:performance}. ``w/o pre." means not using pretrained model. We analyze in turn the three issues about trainset choosing presented in the previous Sec.~\ref{sec:datasetsystem}. For the issue whether choosing train set from the same sensor as the test set helps for generalization, we analyze from the test performance on the SV-hw and SV-sd, although train set SV-bj, SV-all are from the same sensor SuperView as the testset, they are not superior than model trained on US3D-train. The assumption that train set from the same sensor can help generalization is not valid. For the issue whether using diversified regional target distribution can help with generalization, analyzing from the performance of models utilizing SV-bj and SV-all as trainset, SV-bj correlated model generalized better although it only used data from Beijing. Therefore, we can conclude that when collecting same amount of samples for training, directly diversify the target distribution of samples can not help generalization. For the issue whether using similar regional target distribution is helpful for generalization, it can be analyzed from multiple comparative groups. For testset SV-hw, SV-sd, GeoEye-omh, no matter what model structure or training manner we selected, model trained on US3D-train generalizes better than WHU-train, for the reason that WHU-train contains satellite samples from China, while US3D-train contains images of Jacksonville, which has western city architecture style and is more similar to testset in regional target distribution. When it comes to WHU-Stereo-test, models trained on SV-bj and SV-all generalize better than US3D-train, also consistent with our conclusion, revealing the reason that they contain images of Chinese cities, compatible with the regional target distribution of the test set. 

The above analysis suggests that for the dataset, the difference in data characteristics due to the sensors may not have a significant impact after data normalisation applied, but the impact of architectural style is significant. Researchers are advised to choose data with similar regional target distribution or architectural styles to the target dataset for training when facing limited data set resources.
\vspace{-4pt}
\subsection{For training manner}
\begin{figure}[!htp]
  \centering
    \includegraphics[width=3.5in]{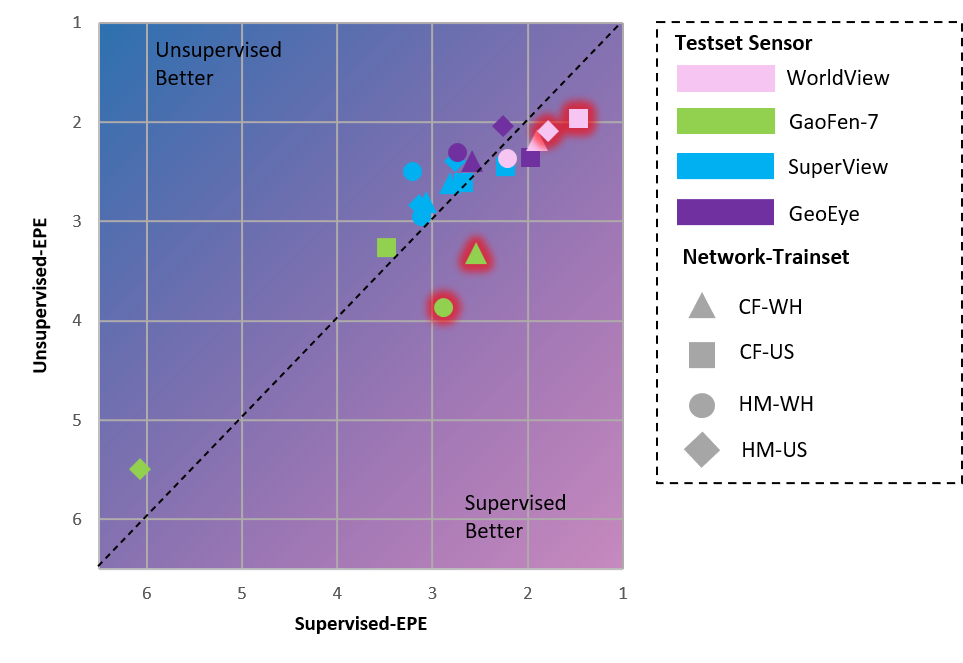}
  \caption{Comparative analysis visualisation of supervised and unsupervised training manners, where the shape of labels denotes network structure and trainset, while color of labels denotes testset. Using the diagonal line as dividing line, the data points closer to unsupervised side represent the corresponding model performs better in unsupervised manner, otherwise similar. Here we mark data points where the trainset and testset are from exactly the same domain as red glowing, which should not be considered in the generalized performance evaluation.}
  \label{fig:unsup_sup}
\end{figure}

\begin{table*}[!htbp]
  \caption{Results about the influence of training manner on generalization performance. To compare the generalization performance between unsupervised and supervised manners in the setting of the same defined network structure and training dataset, we bold the better test results for each pair of generalization tests under the same network structure and train set, and bold the training manner that dominates under this setting. \label{tab:modelperformance}}
  \centering
  \scalebox{1.1}{
\renewcommand{\arraystretch}{1.2} 
    \begin{tabular}{ccccccccccccc}
    \toprule
     \multirow{2}{*}{Model} &\multirow{2}{*}{Trainset} &\multirow{2}{*}{Manner}& \multicolumn{2}{c}{US3D-test} & \multicolumn{2}{c}{WHU-Stereo-test} & \multicolumn{2}{c}{SV-hw} & \multicolumn{2}{c}{SV-sd} &\multicolumn{2}{c}{GeoEye-omh} \\
     \cmidrule(){4-13}
    &&& EPE & D1 & EPE & D1& EPE & D1 & EPE & D1 &EPE & D1\\
    \midrule
    RS-CFNet&US3D-train&\bf{supervised}&1.47&8.23&3.48&33.53&\bf{2.23}&\bf{20.97}&2.67&\bf{22.67}&\bf{1.97}&\bf{15.73}\\
    RS-CFNet&US3D-train&unsupervised&1.96&14.51&\bf{3.26}&\bf{32.31}&2.45&21.83&\bf{2.61}&23.24&2.36&17.02\\
    \hline
    RS-CFNet&WHU-Stereo-train&supervised&\bf{1.90}&\bf{15.24}&2.54&25.69&2.81&24.18&3.06&25.77&2.58&18.88\\
    RS-CFNet&WHU-Stereo-train&\bf{unsupervised}&2.19&19.75&3.32&33.87&\bf{2.62}&\bf{22.46}&\bf{2.81}&\bf{25.04}&\bf{2.40}&\bf{17.73}\\
    \hline
    HMSMNet&US3D-train&supervised&1.78&12.74&6.07&43.78&2.76&25.56&3.14&27.22&2.25&18.42\\
    HMSMNet&US3D-train&\bf{unsupervised}&2.09&18.34&\bf{5.50}&\bf{38.63}&\bf{2.40}&\bf{23.29}&\bf{2.84}&\bf{25.72}&\bf{2.04}&\bf{17.73}\\
    \hline
    HMSMNet&WHU-Stereo-train&supervised&\bf{2.21}&\bf{19.98}&2.89&29.13&3.21&29.67&3.12&28.07&2.74&23.34\\
    HMSMNet&WHU-Stereo-train&\bf{unsupervised}&2.37&23.13&3.87&36.73&\bf{2.50}&\bf{24.16}&\bf{2.95}&\bf{26.66}&\bf{2.31}&\bf{20.16}\\
 
     \bottomrule
 \end{tabular}}
 \label{tab:mannerperformance}
  \end{table*}
 \begin{table*}[!htb]
  \caption{Results about the influence of training setting on generalization performance. To compare the effect of the training setting on performance and generalization performance, we fix the network structure and trainset to compare its effect with and without the early-stop consistency criterion, denoted as `w\ crit.' and `w/o\ crit.'. We bold the better results and better training setting.\label{tab:trainingsetting}}
  \centering
  \scalebox{1.0}{
\renewcommand{\arraystretch}{1.1} 
    \begin{tabular}{ccccccccccccc}
    \toprule
     \multirow{2}{*}{Model} & \multirow{2}{*}{Trainset} & \multirow{2}{*}{Manner} &\multicolumn{2}{c}{US3D-test} & \multicolumn{2}{c}{WHU-Stereo-test} & \multicolumn{2}{c}{SV-hw} & \multicolumn{2}{c}{SV-sd} &\multicolumn{2}{c}{GeoEye-omh} \\
     \cmidrule(){4-13}
    &&& EPE & D1 & EPE & D1& EPE & D1 & EPE & D1 &EPE & D1\\
    \midrule
    RS-CFNet&US3D-train&unsupervised-w/o\ crit.&2.08&18.30&\bf{3.15}&33.80&\bf{2.44}&22.47&2.80&25.58&\bf{2.14}&17.90\\
    RS-CFNet&US3D-train&\bf{unsupervised-w\ crit.}&\bf{1.96}&\bf{14.51}&3.26&\bf{32.31}&2.45&\bf{21.83}&\bf{2.61}&\bf{23.24}&2.36&\bf{17.02}\\
    \hline
    RS-CFNet&WHU-Stereo-train&unsupervised-w/o\ crit.&2.26&20.76&3.47&34.16&2.67&22.88&2.84&25.45&\bf{2.31}&18.83\\
    RS-CFNet&WHU-Stereo-train&\bf{unsupervised-w\ crit.}&\bf{2.19}&\bf{19.75}&\bf{3.32}&\bf{33.87}&\bf{2.62}&\bf{22.46}&\bf{2.81}&\bf{25.04}&2.40&\bf{17.73}\\
     \bottomrule
 \end{tabular}}
 \label{tab:settingperformance}
  \end{table*}
Respectively utilizing network RS-CFNet and HMSMNet, for both supervised and unsupervised training manner, we conduct sufficient experiments to explore the effect of the training method on generalization ability, the results of which are presented in Tab.~\ref{tab:mannerperformance}. As shown in Fig.~\ref{fig:unsup_sup}, we use different colors to represent sensors from which the test set comes, and different shapes to represent the network structure and training set. The horizontal and vertical coordinates represent the EPE of the supervised method and the EPE of the unsupervised method, respectively. We inverted the axes to clearly show the effectiveness comparison of the unsupervised and supervised methods with the same network structure, the same training set, and the same test set.
Clearly, except for the models tested on exactly corresponding sensor data, which we marked with red boxes, 75\% of the comparison examples have better unsupervised results than supervised test results, which can indicate that the unsupervised manner has stronger generalization ability than the supervised manner with same network structure and same training dataset. Researchers are suggested that when it is not possible to get train set from the same sensors and similar ground area as testset, it is more likely to get good generalization by using unsupervised method for training, instead of supervised method.
  
 \begin{figure*}[!ht]
    \centering
    \captionsetup{justification=centering}
    \begin{subfigure}[]{0.39\columnwidth}
        \rotatebox{90}{\scriptsize{~~~~~~~~~~~~Left Image}}
        \begin{minipage}[t]{0.9\linewidth}
        \includegraphics[width=1\columnwidth]{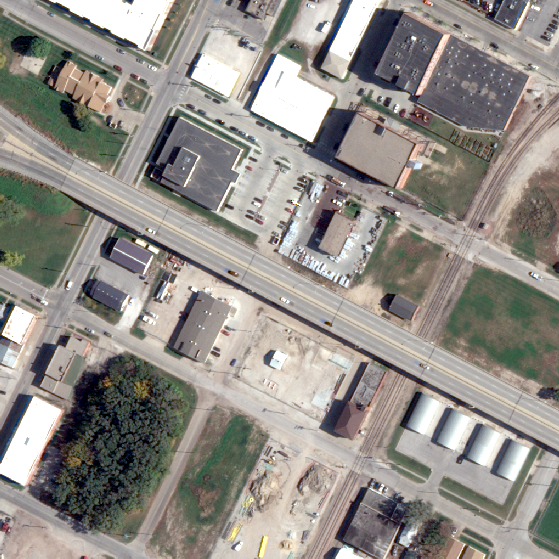}
        \end{minipage}
    \end{subfigure}
    \begin{subfigure}[]{0.35\columnwidth}
        \includegraphics[width=1\columnwidth]{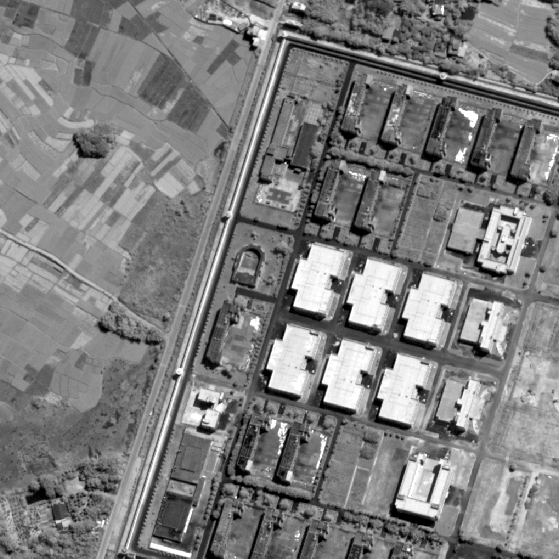}
    \end{subfigure}
    \begin{subfigure}[]{0.36\columnwidth}
        \includegraphics[width=1\columnwidth]{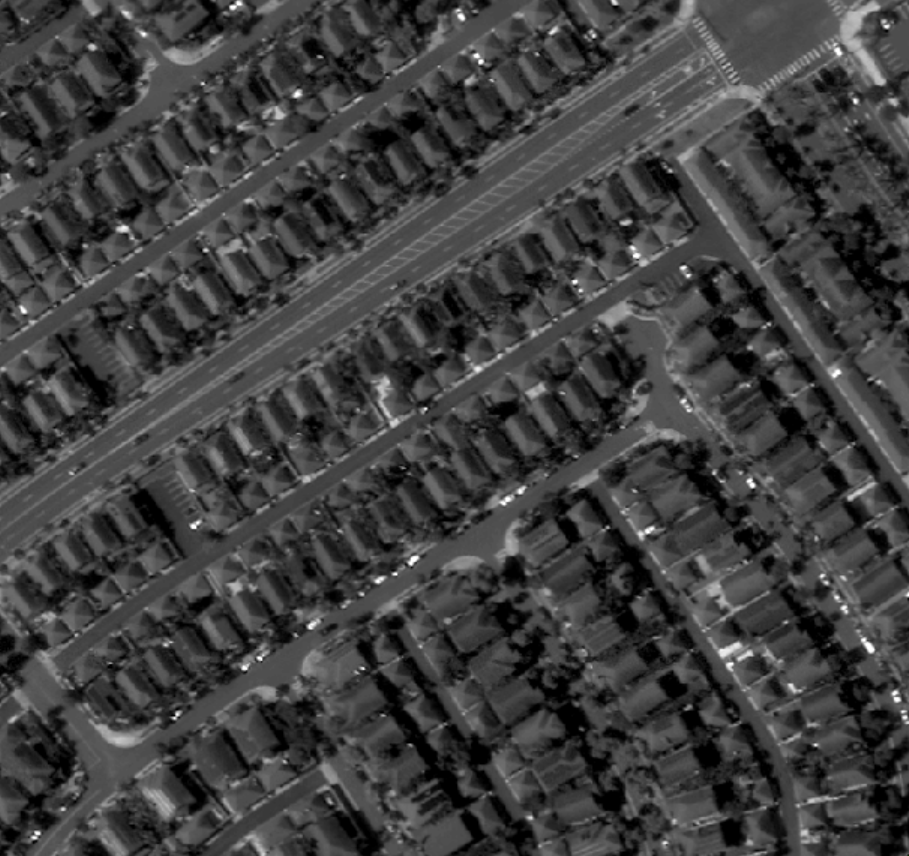}
    \end{subfigure}
    \begin{subfigure}[]{0.35\columnwidth}
        \includegraphics[width=1\columnwidth]{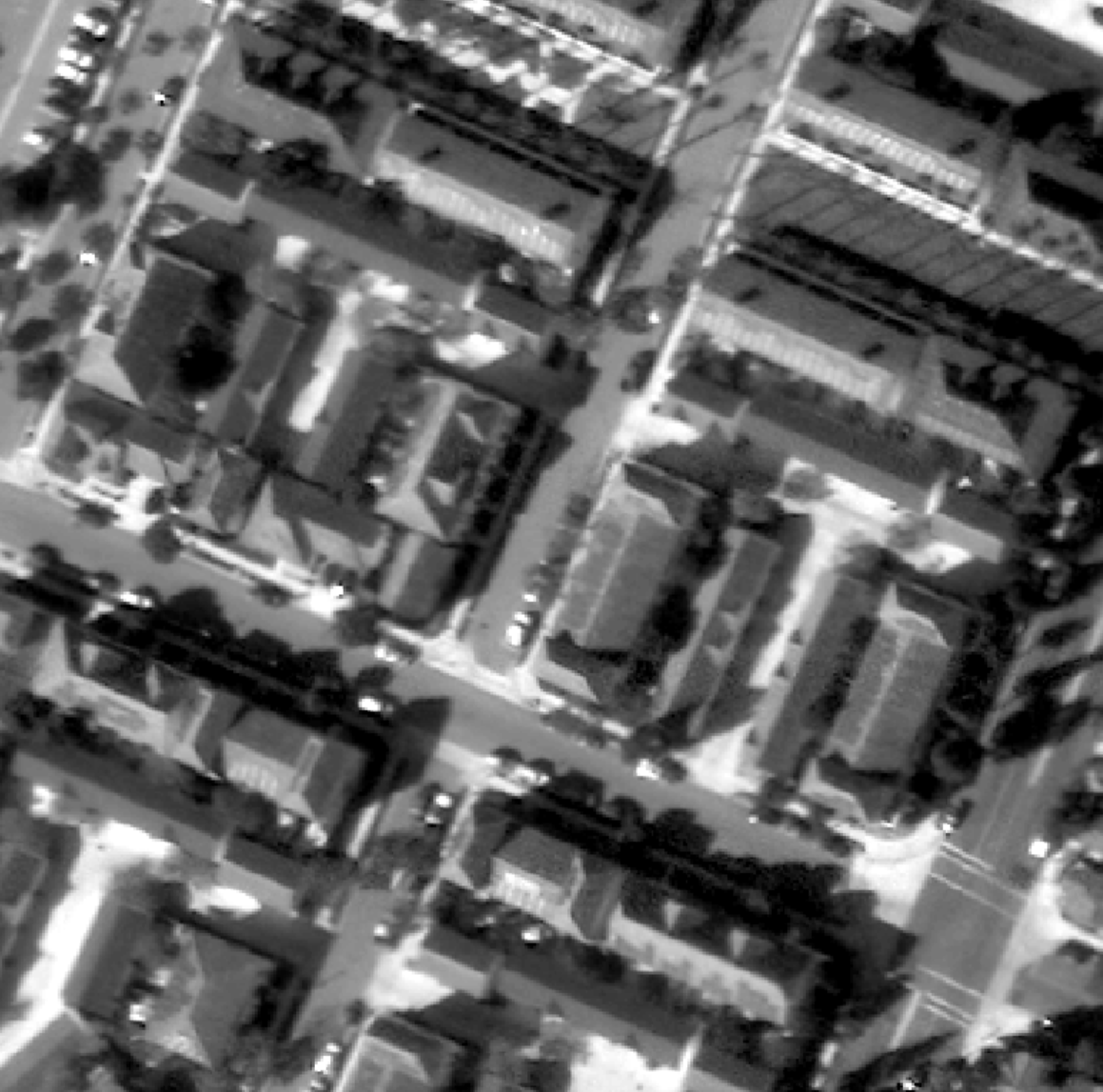}
    \end{subfigure}
    \begin{subfigure}[]{0.33\columnwidth}
        \includegraphics[width=1\columnwidth]{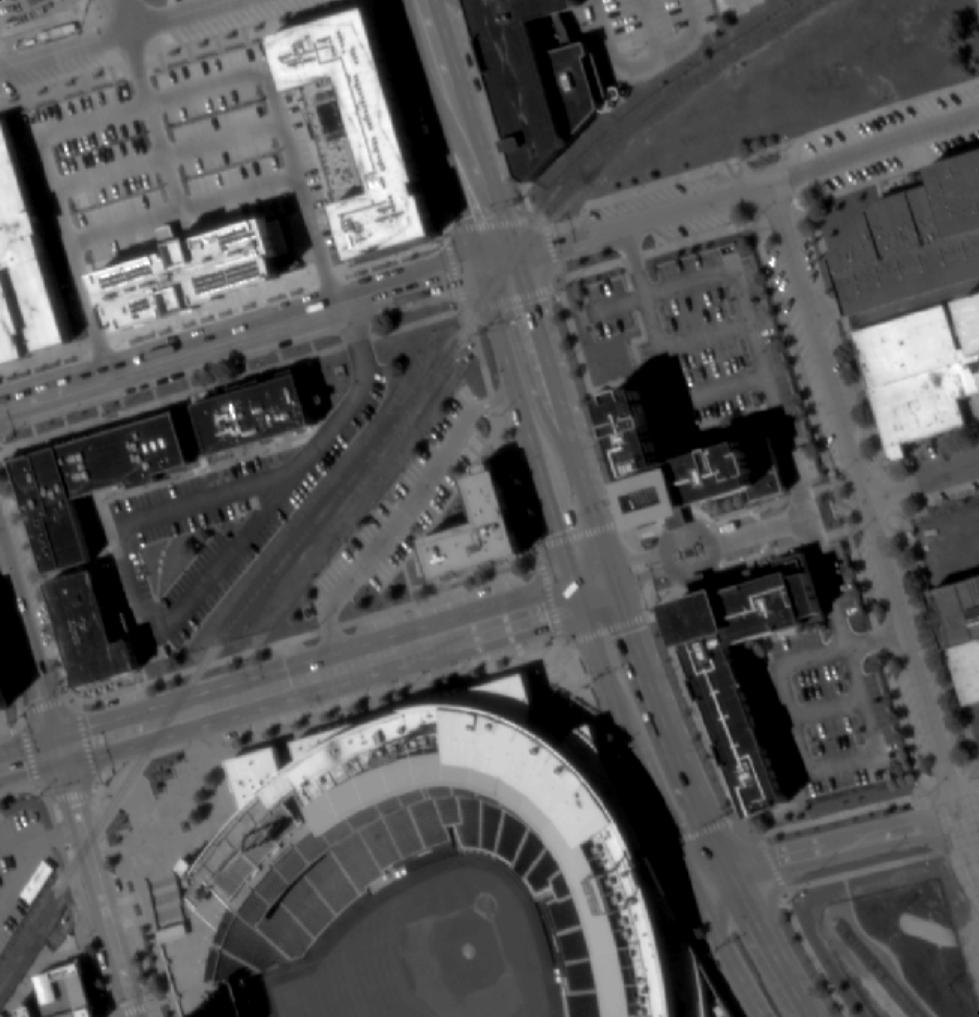}
    \end{subfigure}
    \\
    \centering
    \begin{subfigure}[]{0.39\columnwidth}
    \rotatebox{90}{\scriptsize{~~~~~~~~~~~Ground~Truth}}
        \begin{minipage}[t]{0.9\linewidth}
        \includegraphics[width=1\columnwidth]{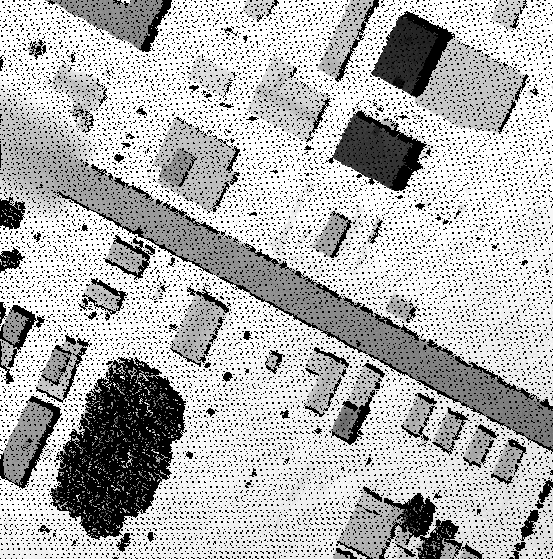}
        \end{minipage}
    \end{subfigure}
    \begin{subfigure}[]{0.35\columnwidth}
        \includegraphics[width=1\columnwidth]{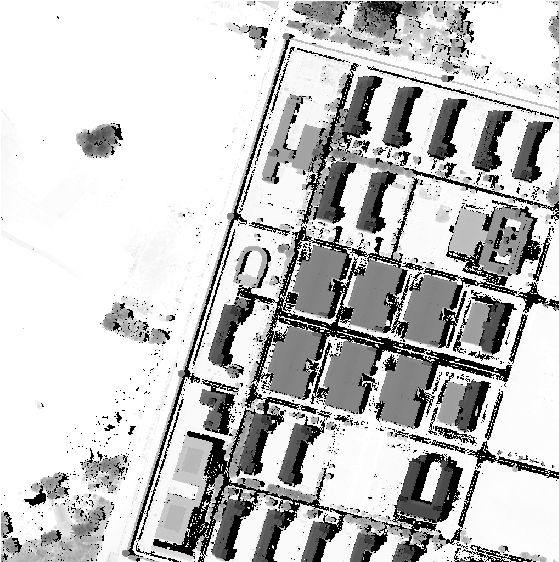}
    \end{subfigure}
    \begin{subfigure}[]{0.36\columnwidth}
        \includegraphics[width=1\columnwidth]{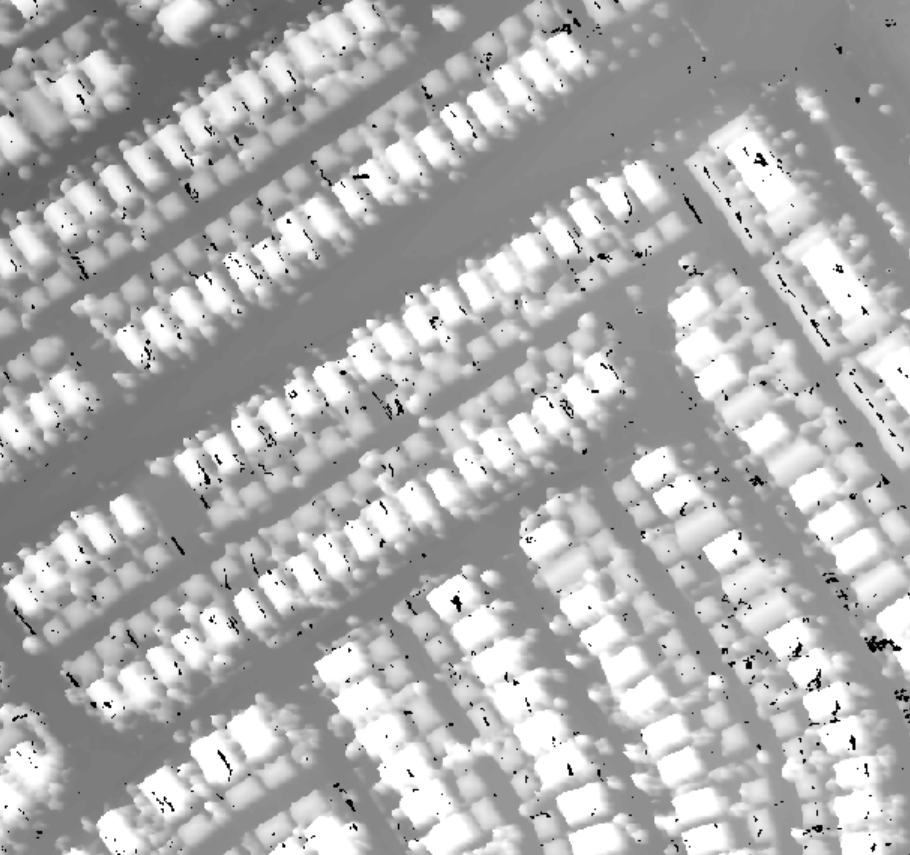}
    \end{subfigure}
    \begin{subfigure}[]{0.35\columnwidth}
        \includegraphics[width=1\columnwidth]{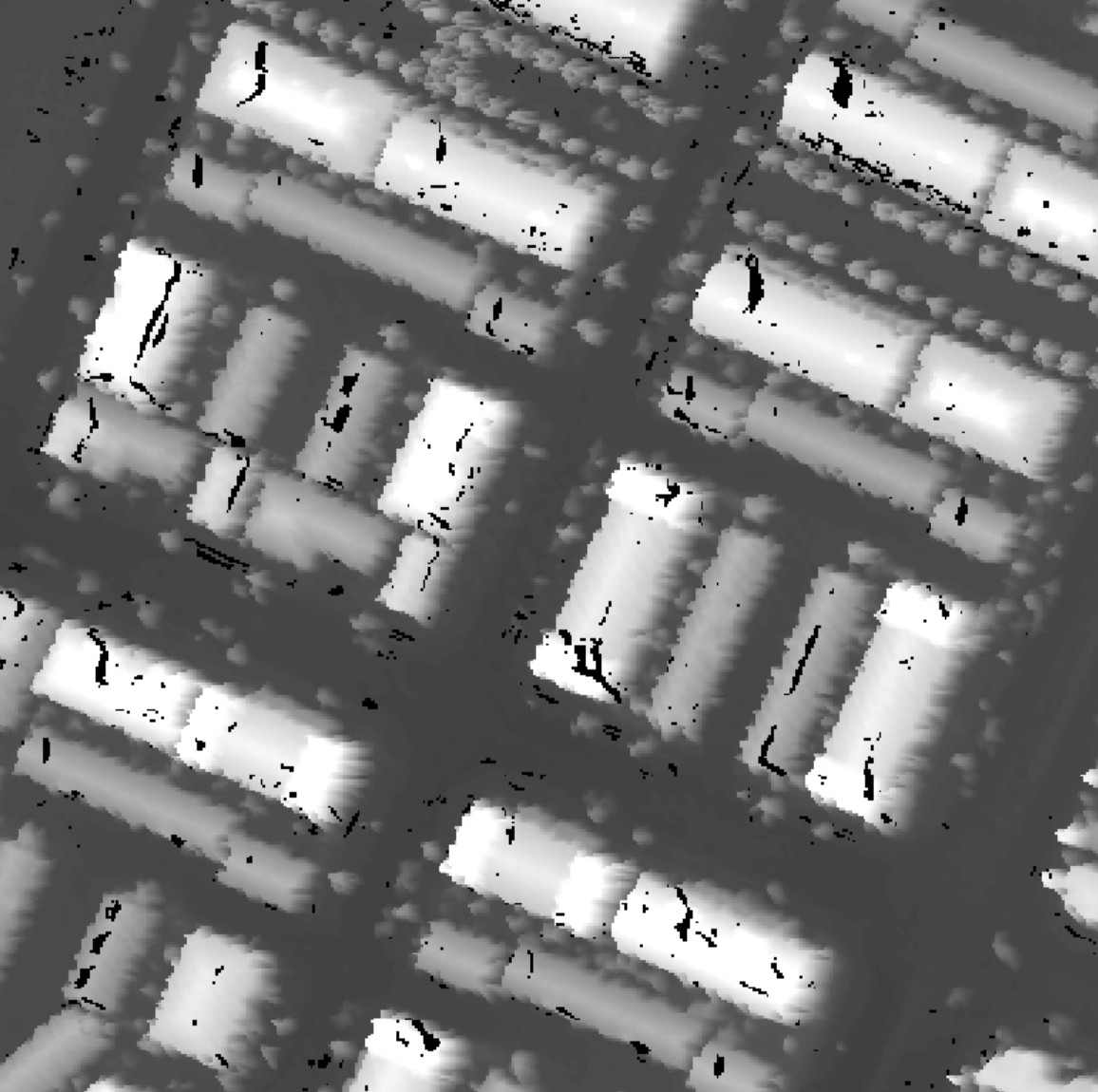}
    \end{subfigure}
    \begin{subfigure}[]{0.33\columnwidth}
        \includegraphics[width=1\columnwidth]{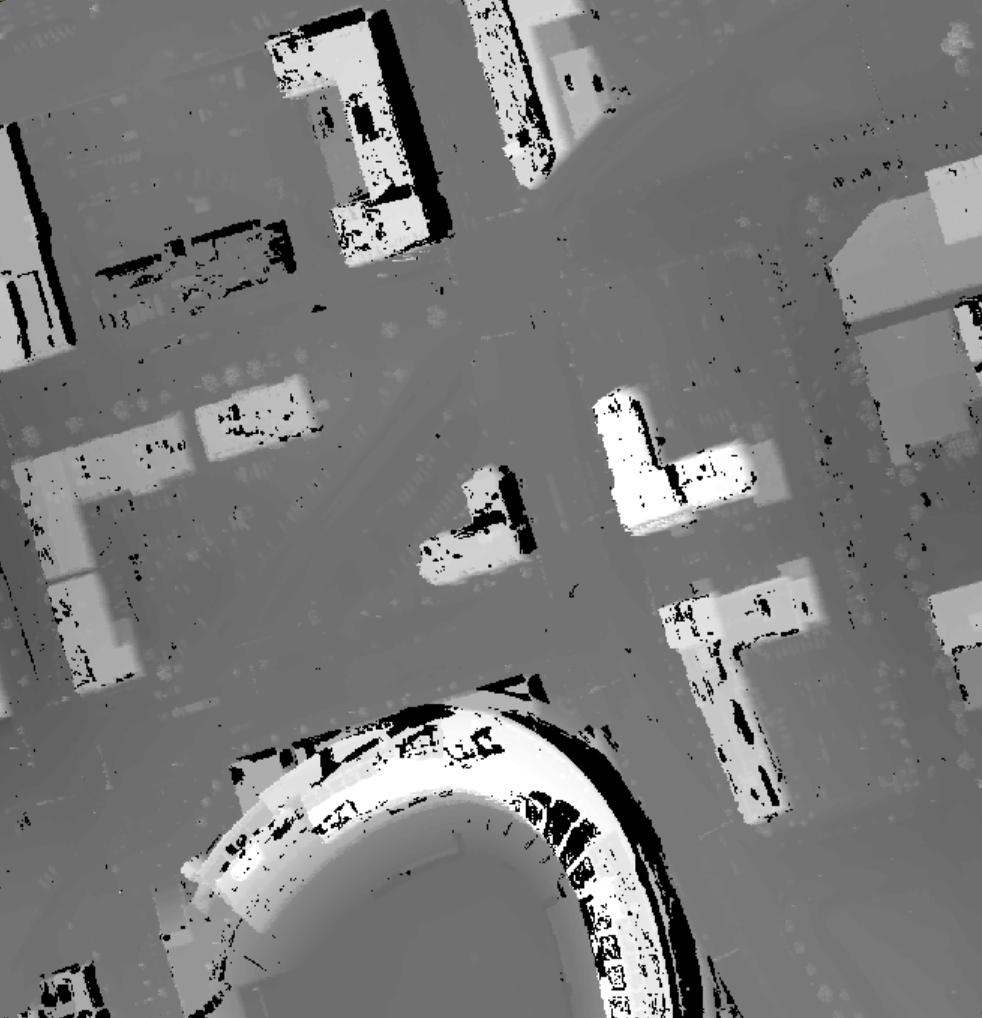}
    \end{subfigure}
    \centering
    \\
    \begin{subfigure}[]{0.39\columnwidth}
        \centering
        \rotatebox{90}{\scriptsize{~~~~~~~~WHU-w/o crit.}}
        \begin{minipage}[t]{0.9\linewidth}
        \includegraphics[width=1\columnwidth]{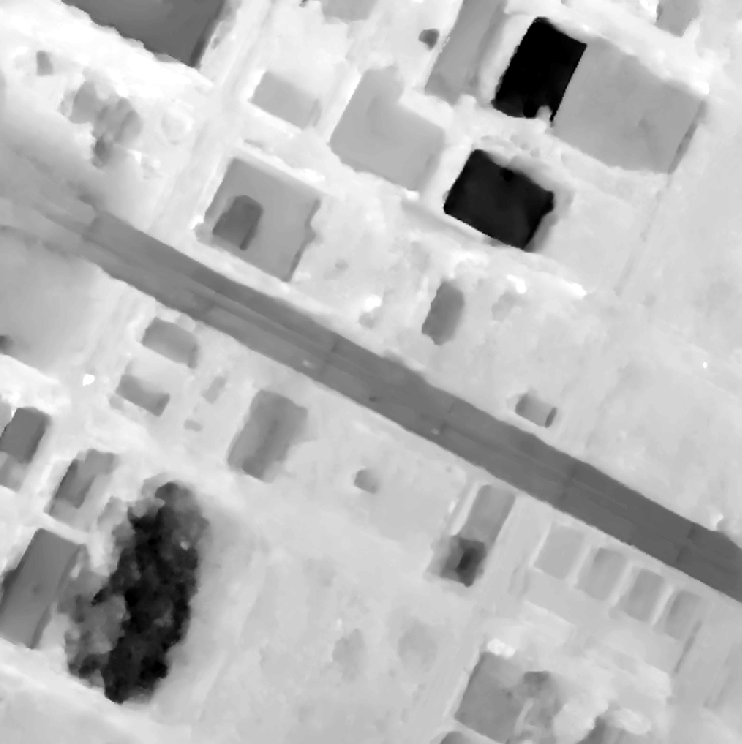}
        \end{minipage}
    \end{subfigure}
    \begin{subfigure}[]{0.35\columnwidth}
        \includegraphics[width=1\columnwidth]{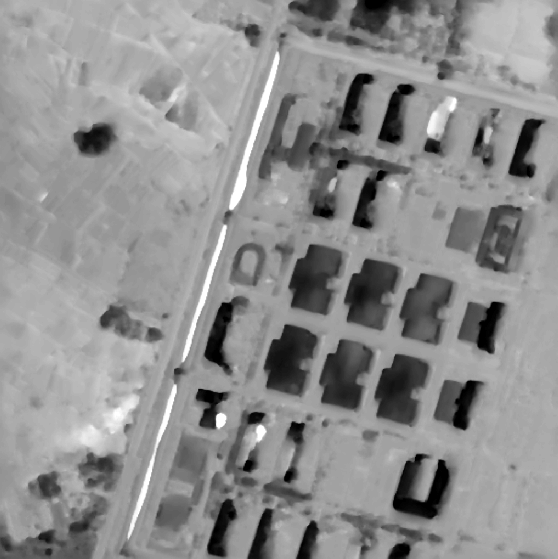}
    \end{subfigure}
    \begin{subfigure}[]{0.36\columnwidth}
        \includegraphics[width=1\columnwidth]{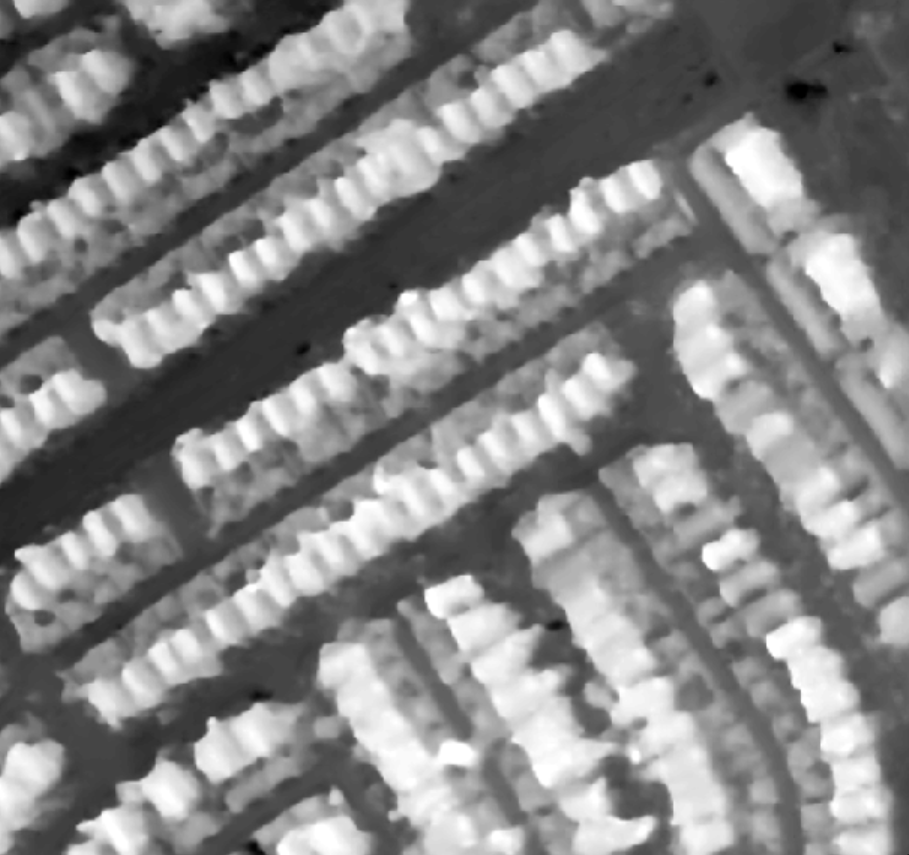}
    \end{subfigure}
    \begin{subfigure}[]{0.35\columnwidth}
        \includegraphics[width=1\columnwidth]{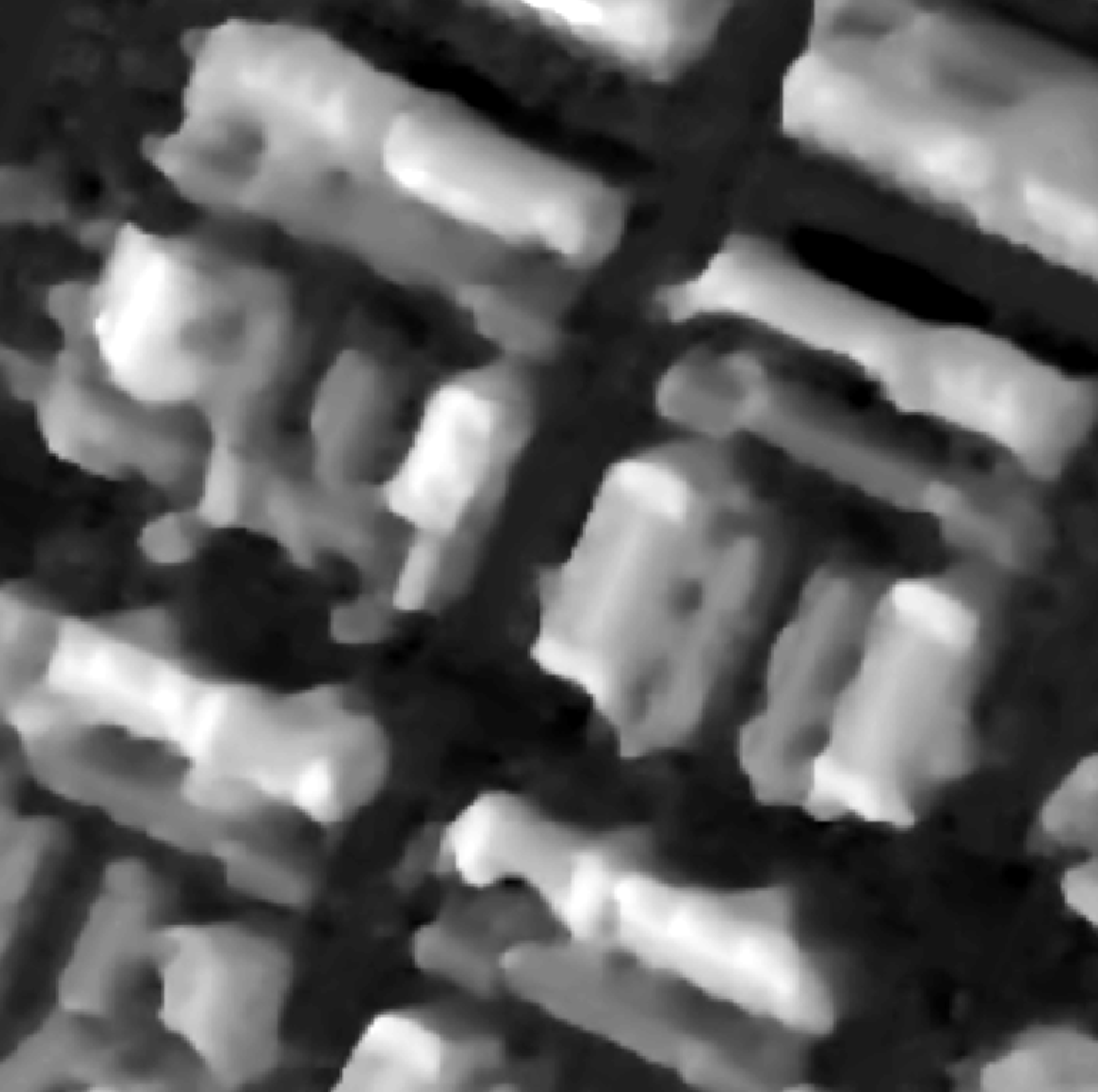}
    \end{subfigure}
    \begin{subfigure}[]{0.33\columnwidth}
        \includegraphics[width=1\columnwidth]{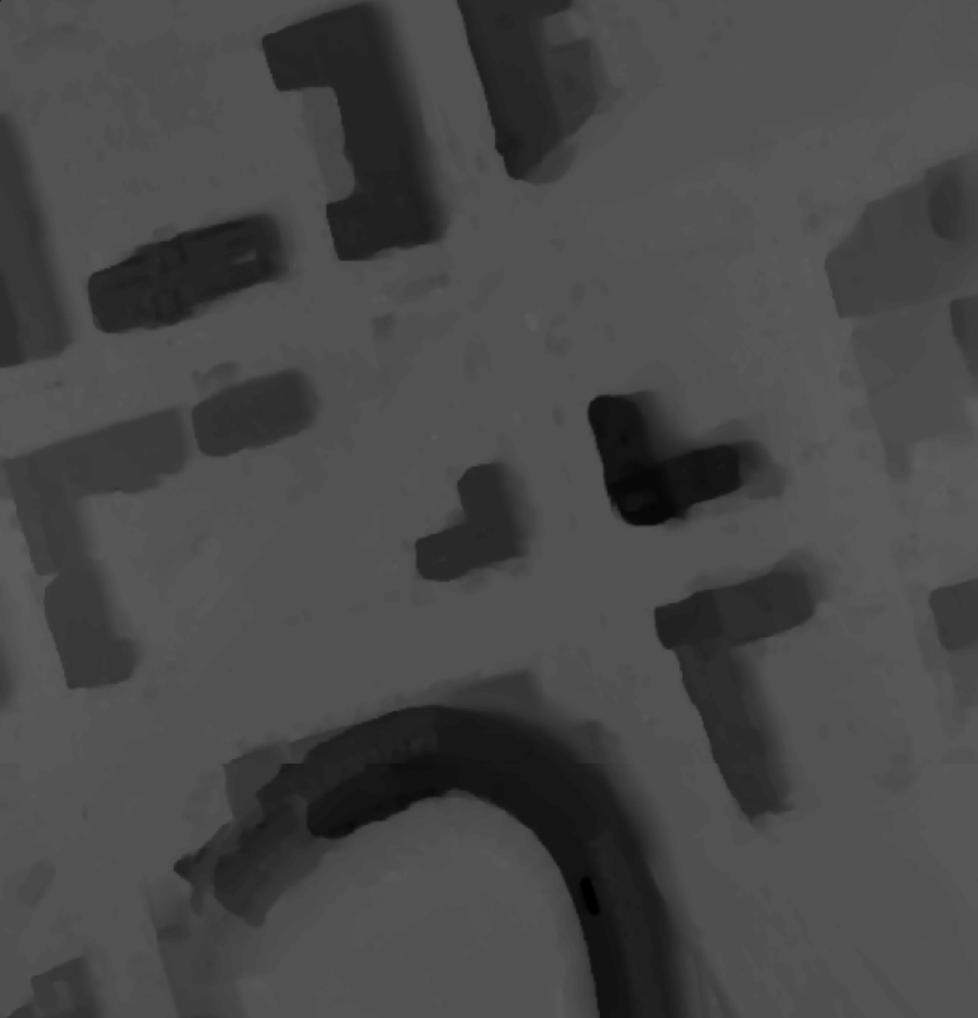}
    \end{subfigure}
    \\
    \begin{subfigure}[]{0.39\columnwidth}
        \centering
        \rotatebox{90}{\scriptsize{~~~~~~~~~~~WHU-w crit.}}
        \begin{minipage}[t]{0.9\linewidth}
        \includegraphics[width=1\columnwidth]{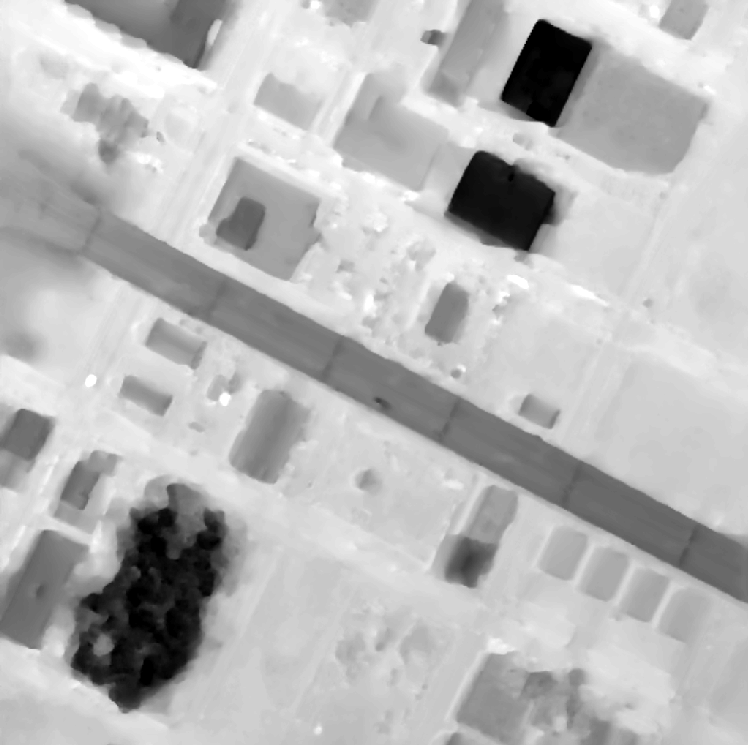}
        \end{minipage}
    \end{subfigure}
    \begin{subfigure}[]{0.35\columnwidth}
        \includegraphics[width=1\columnwidth]{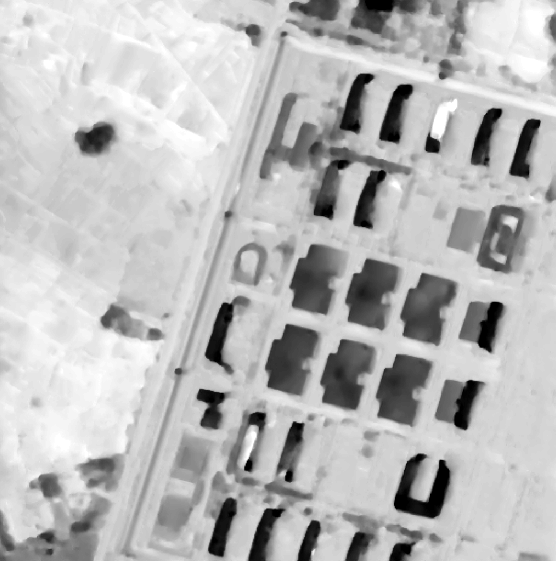}
    \end{subfigure}
    \begin{subfigure}[]{0.36\columnwidth}
        \includegraphics[width=1\columnwidth]{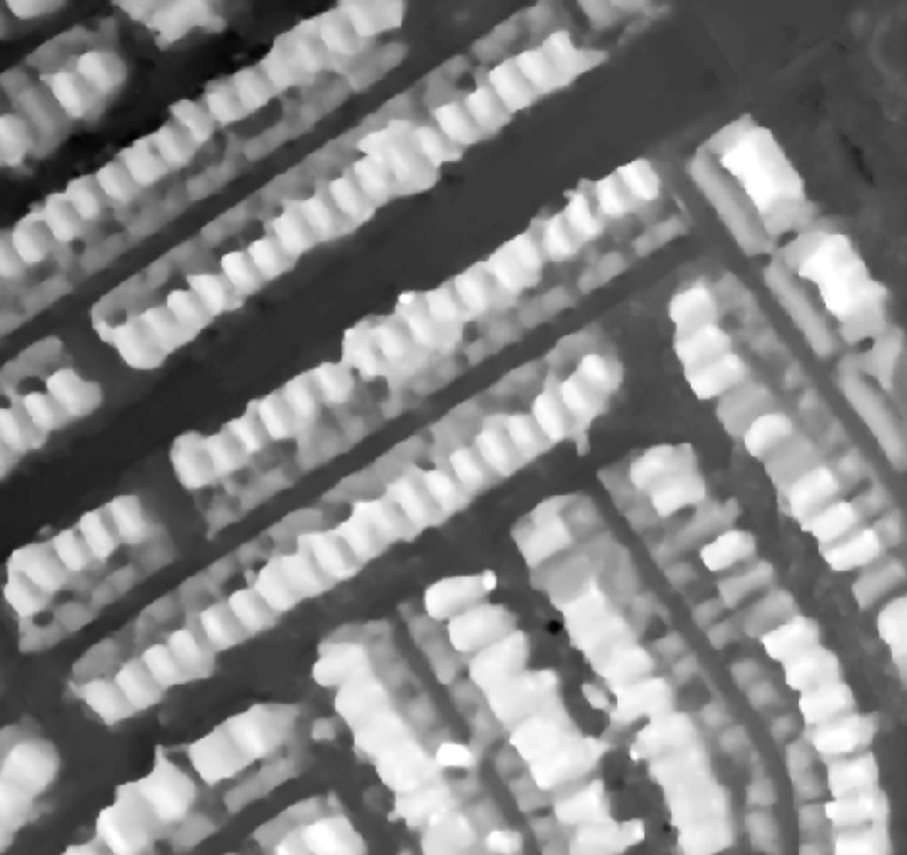}
    \end{subfigure}
    \begin{subfigure}[]{0.35\columnwidth}
        \includegraphics[width=1\columnwidth]{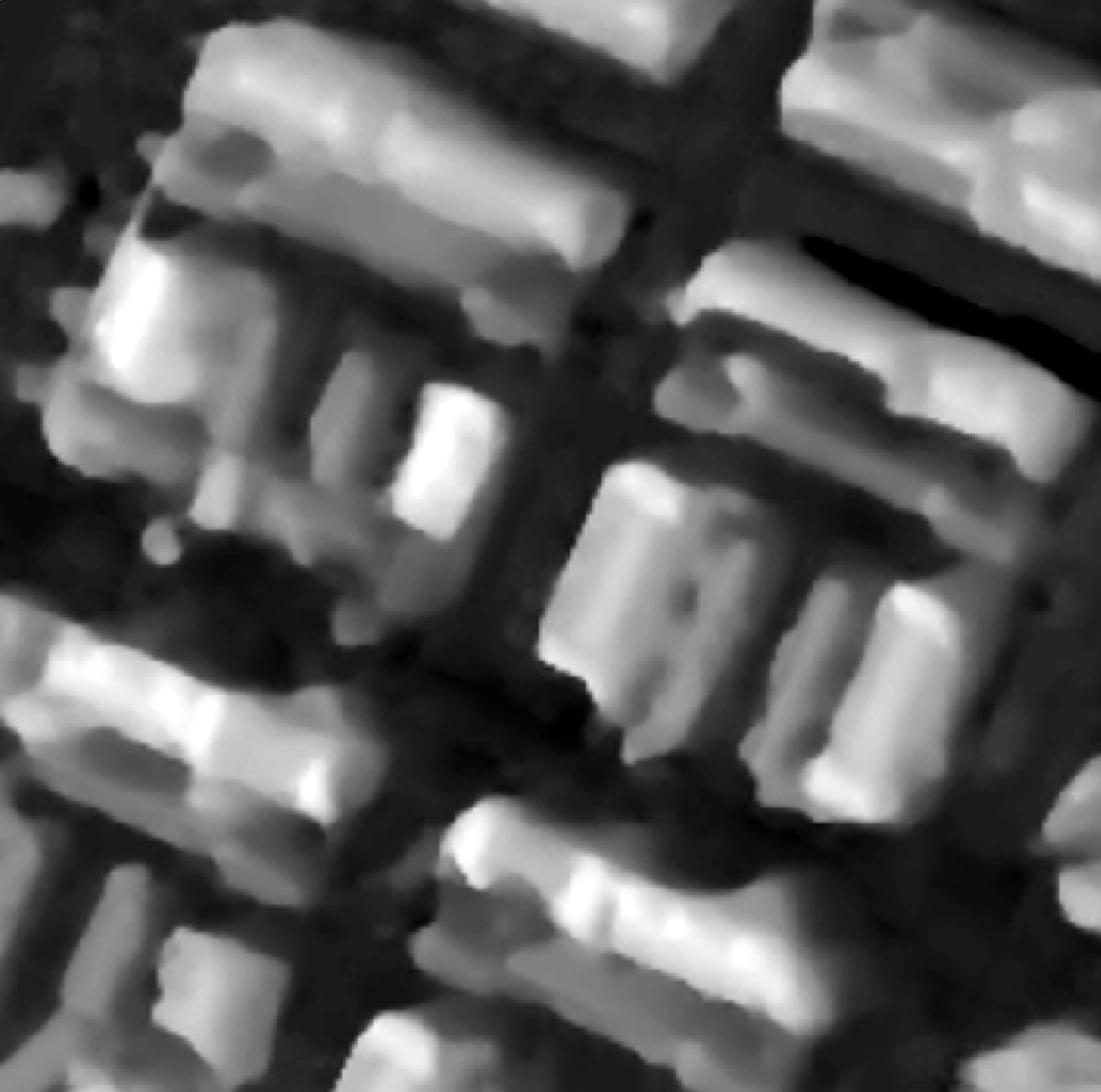}
    \end{subfigure}
    \begin{subfigure}[]{0.33\columnwidth}
        \includegraphics[width=1\columnwidth]{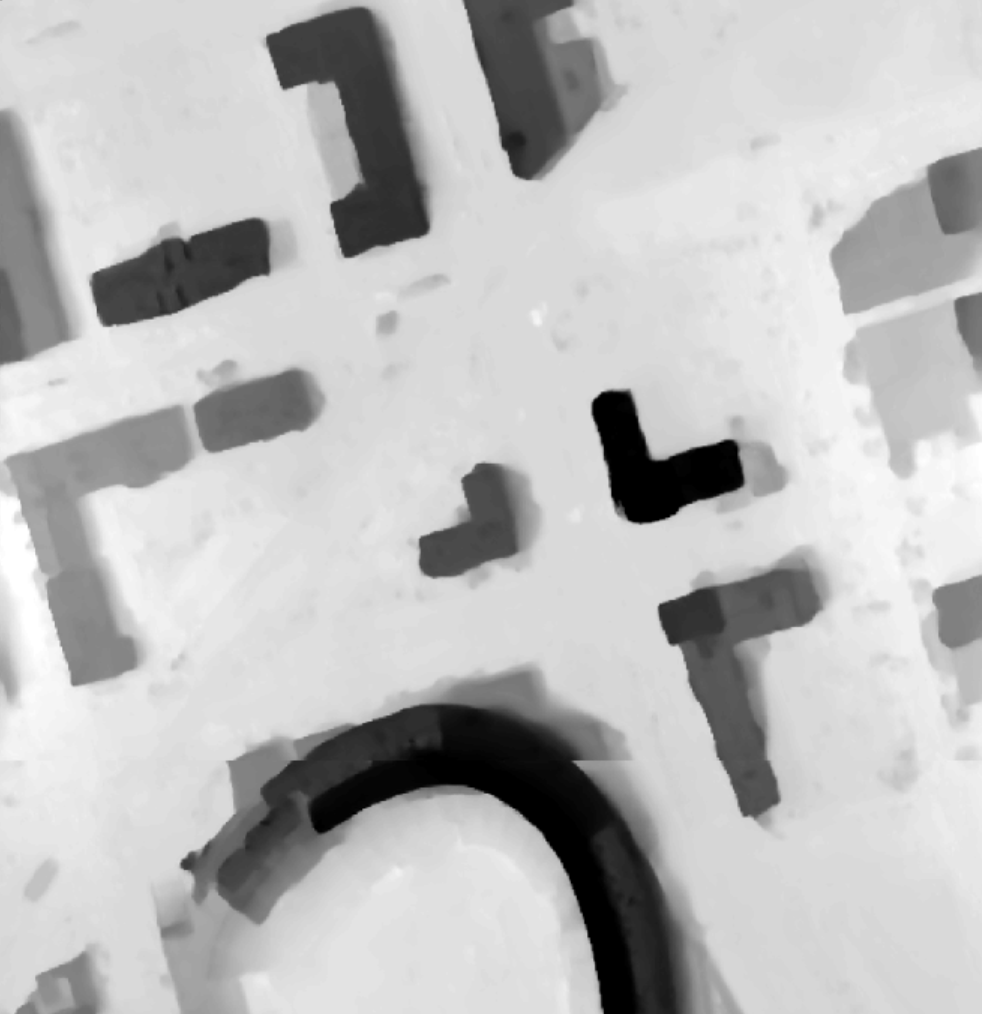}
    \end{subfigure}
    \\
    \centering
    \begin{subfigure}[]{0.39\columnwidth}
    \rotatebox{90}{\scriptsize{~~~~~~~~~US3D-w/o crit.}}
        \begin{minipage}[t]{0.9\linewidth}
        \includegraphics[width=1\columnwidth]{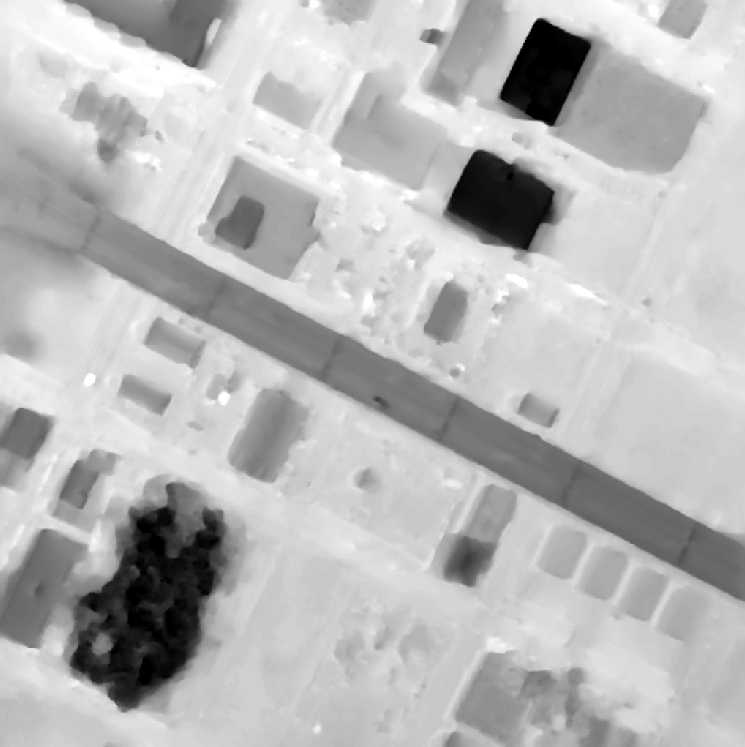}
        \end{minipage}
    \end{subfigure}
    \begin{subfigure}[]{0.35\columnwidth}
        \includegraphics[width=1\columnwidth]{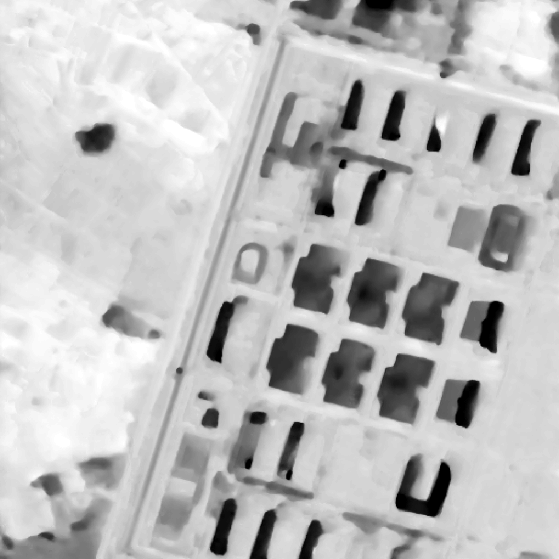}
    \end{subfigure}
    \begin{subfigure}[]{0.36\columnwidth}
        \includegraphics[width=1\columnwidth]{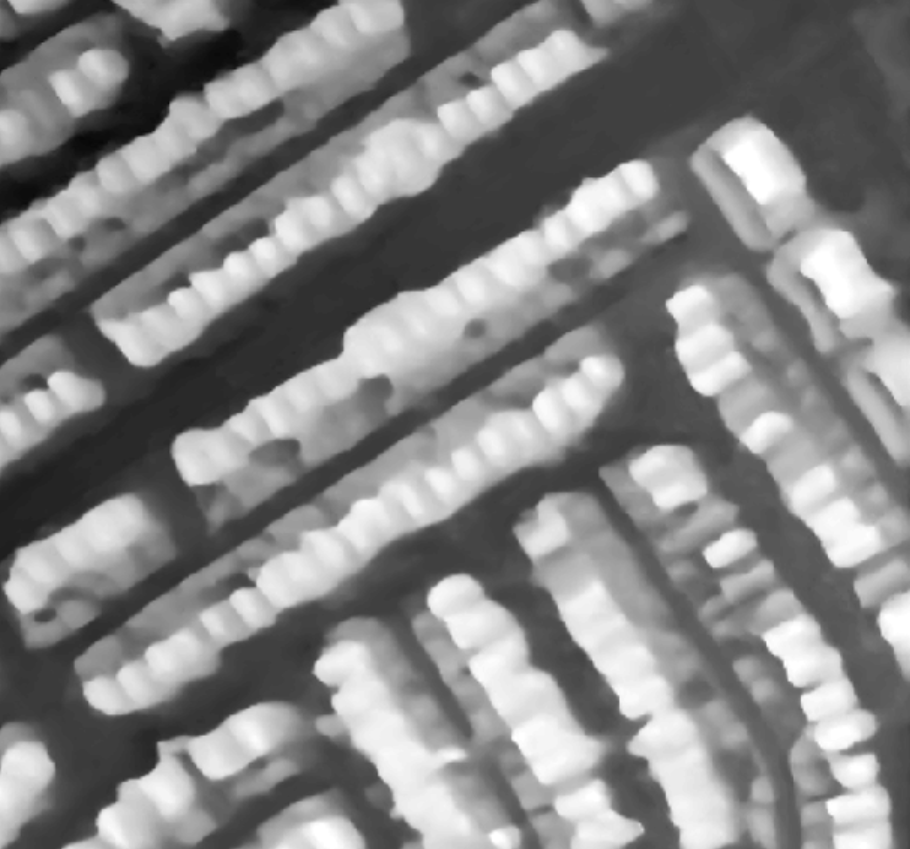}
    \end{subfigure}
    \begin{subfigure}[]{0.35\columnwidth}
        \includegraphics[width=1\columnwidth]{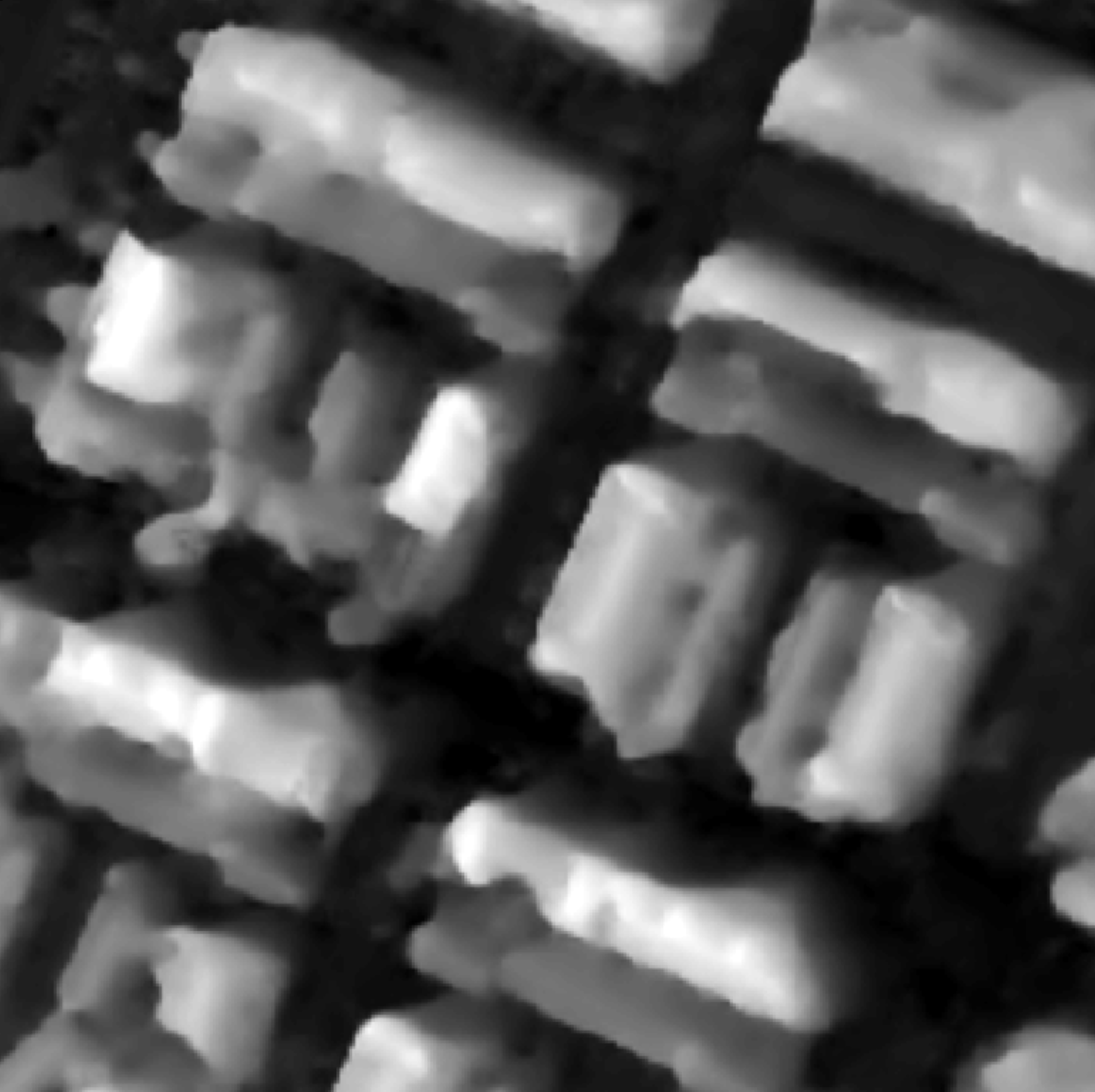}
    \end{subfigure}
    \begin{subfigure}[]{0.33\columnwidth}
        \includegraphics[width=1\columnwidth]{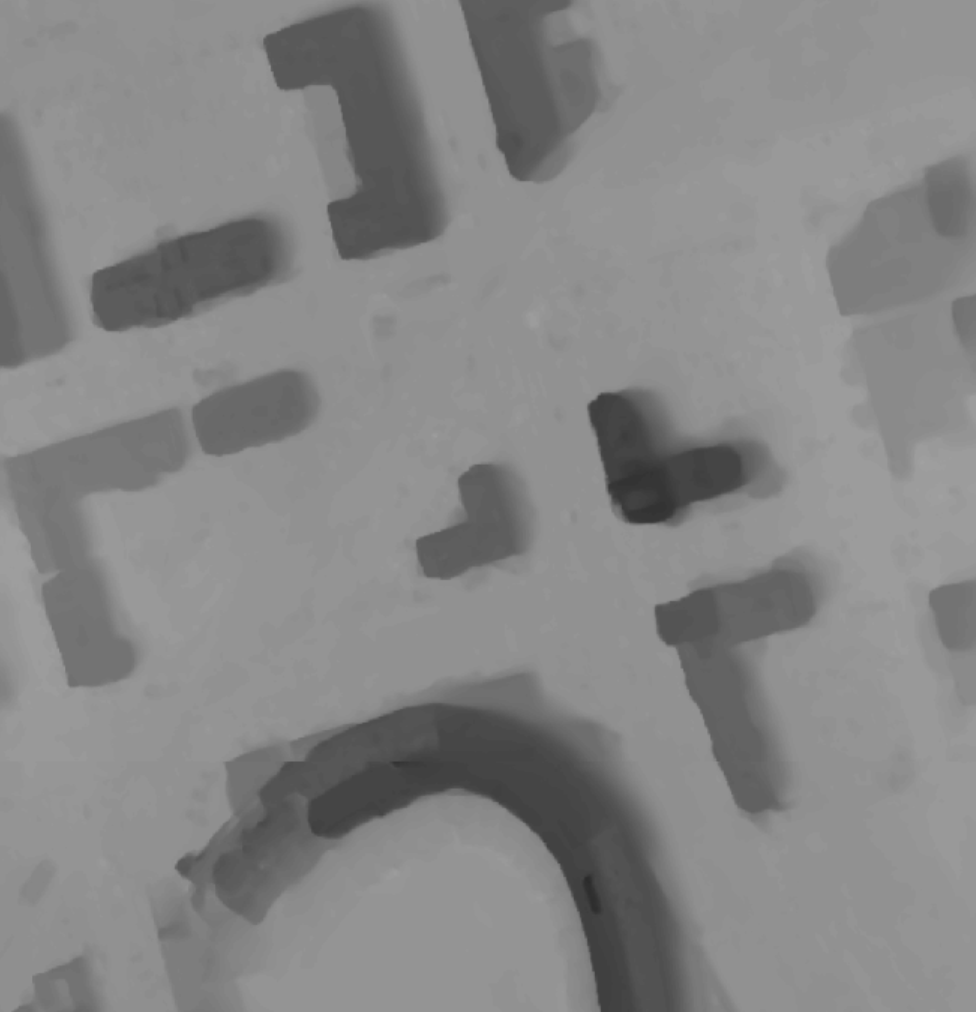}
    \end{subfigure}
    \centering
    \\
    
	\setcounter{subfigure}{0}
    \begin{subfigure}[US3D-test]{0.39\columnwidth}
        \centering
        \rotatebox{90}{\scriptsize{~~~~~~~~~~US3D-w crit.}}
        \begin{minipage}[t]{0.9\linewidth}
        \includegraphics[width=1\columnwidth]{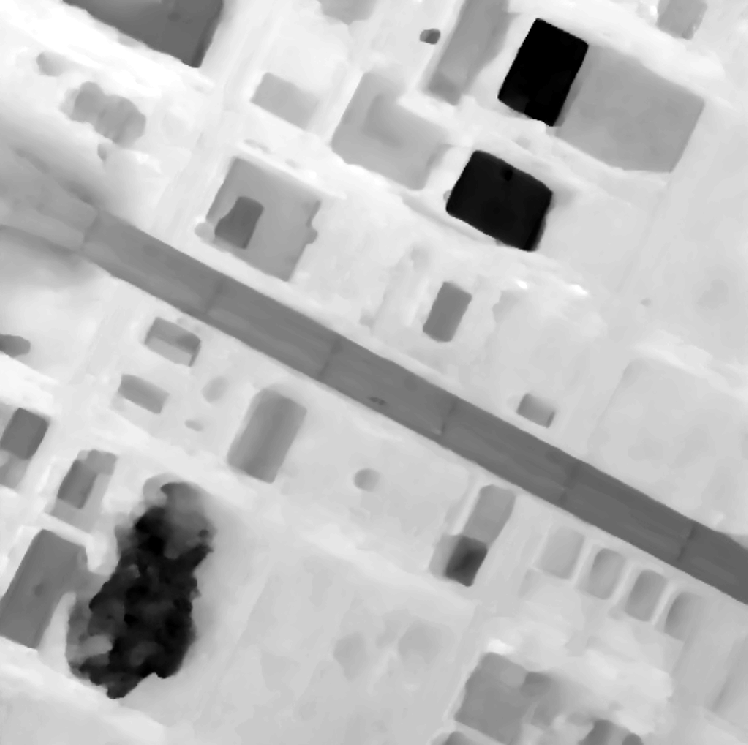}
        \end{minipage}
        \vspace{-10pt}
        \caption{US3D-test}
    \end{subfigure}
    \begin{subfigure}[WHU-Stereo-test]{0.35\columnwidth}
        \includegraphics[width=1\columnwidth]{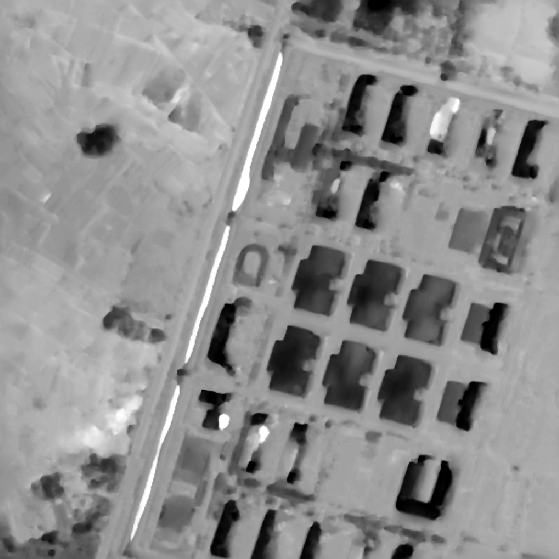}
        \vspace{-10pt}
        \caption{WHU-Stereo-test}
    \end{subfigure}
    \begin{subfigure}[SV-hw]{0.36\columnwidth}
        \includegraphics[width=1\columnwidth]{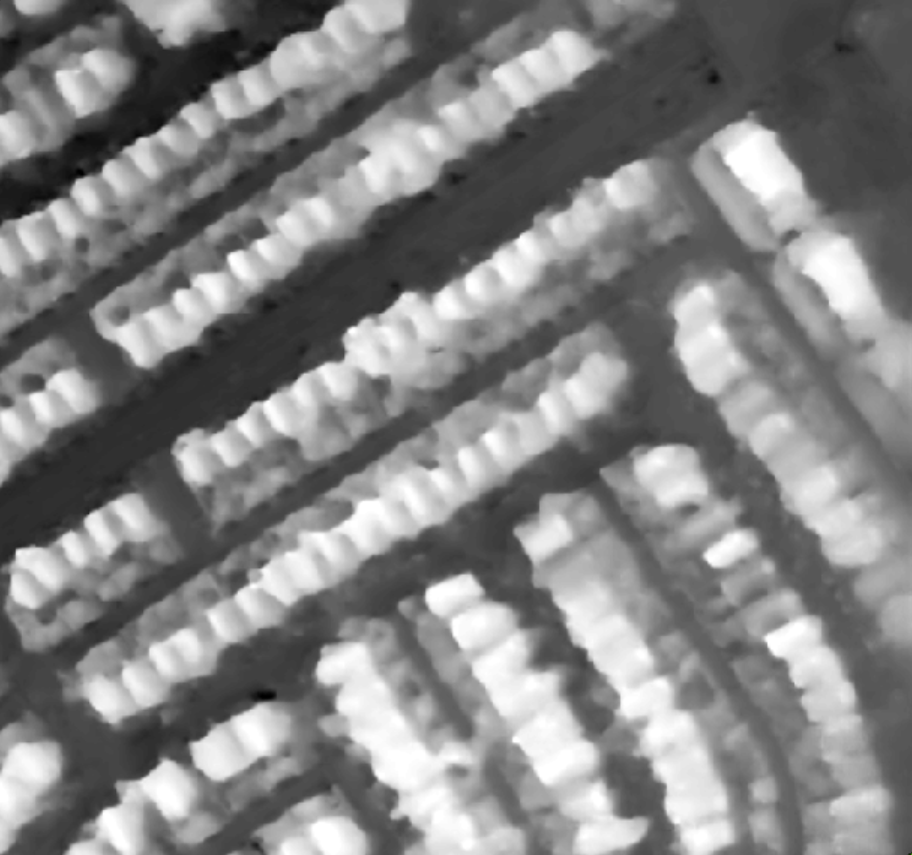}
        \vspace{-10pt}
        \caption{SV-hw}
    \end{subfigure}
    \begin{subfigure}[SV-sd]{0.35\columnwidth}
        \includegraphics[width=1\columnwidth]{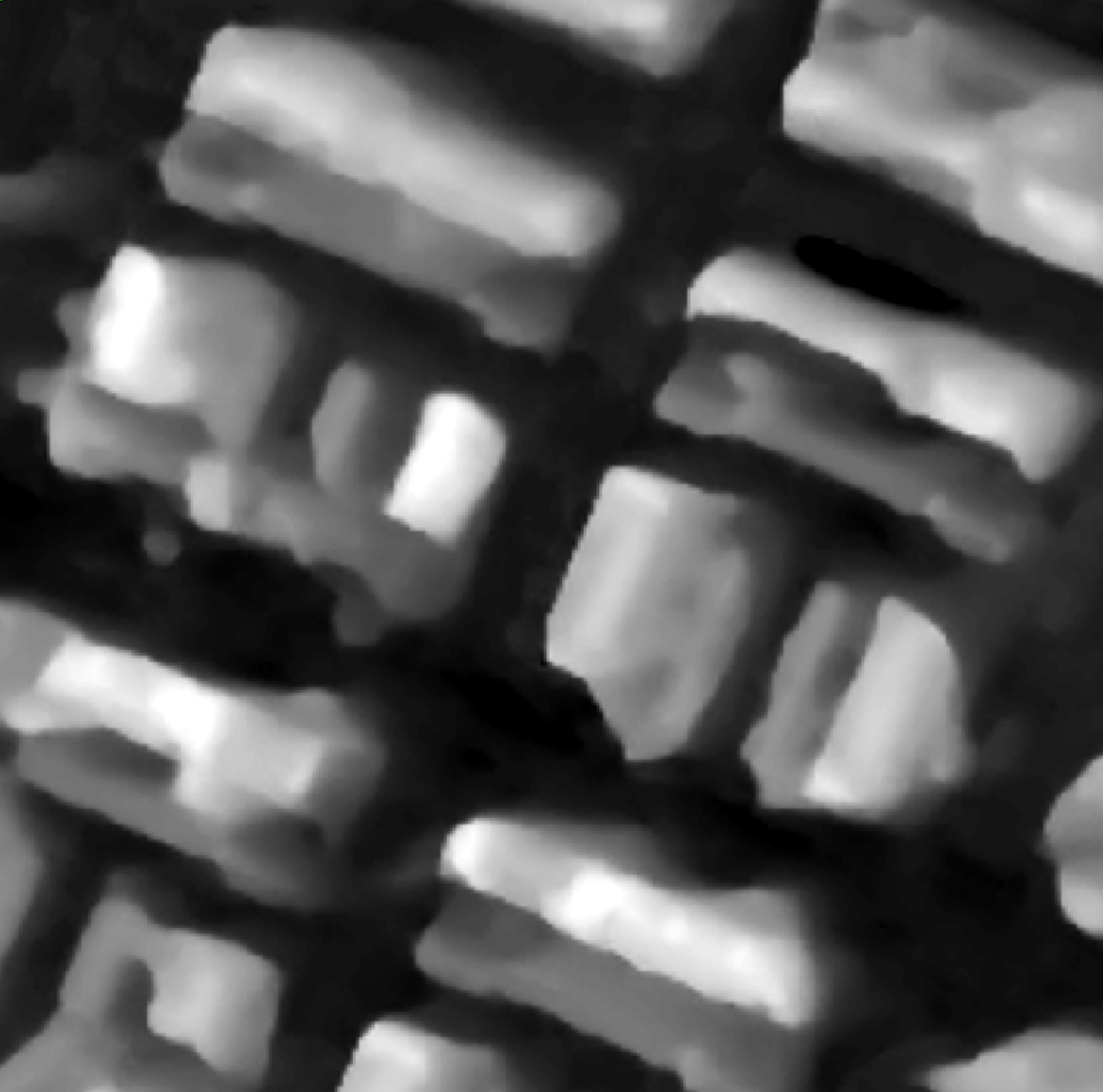}
        \vspace{-10pt}
        \caption{SV-sd}
    \end{subfigure}
    \begin{subfigure}[SV-omh]{0.33\columnwidth}
        \includegraphics[width=1\columnwidth]{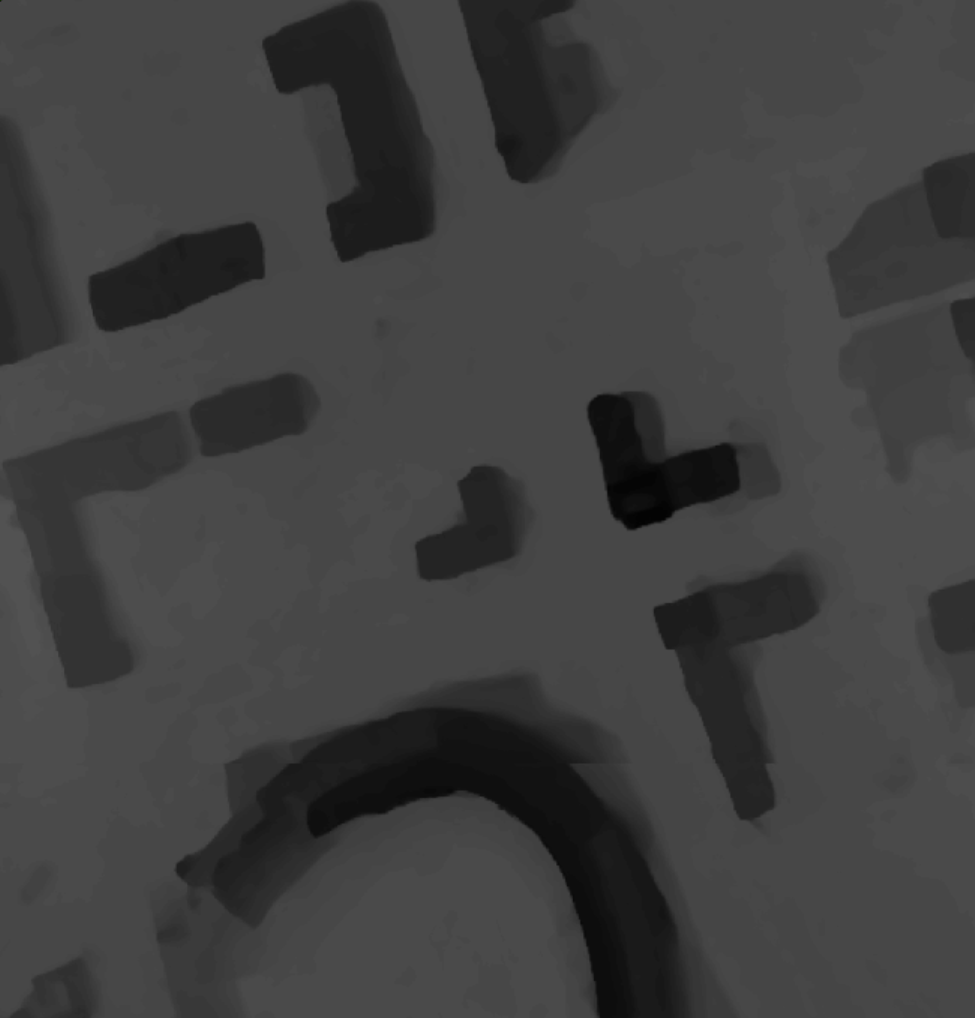}
        \vspace{-10pt}
        \caption{SV-omh}
    \end{subfigure}
\vspace{-7pt}
    \caption{The visualization comparison of generalization performance of models trained with(unsupervised-w crit.) and without(unsupervised-w/o crit.) early-stop consistency criterion. For both models trained with WHU-Stereo-train and US3D-train, early-stop strategy helps to gain better and stable generalization ability on testsets.}
    \label{fig:trainingsetting}
    \vspace{-20pt}
\end{figure*}
\vspace{-7pt}
\vspace{-5pt}
\subsection{For training setting}
\vspace{-3pt}
To verify the reliability of our proposed early-stop consistency criterion, we conducted another comparison experiment. It has been preliminarily verified earlier on the dataset US3D and WHU-Stereo with ground truth that the criterion indeed stops the model at the appropriate iteration to improve the performance, as shown in Sec.~\ref{earlystop}. Here we analyze the generalization performance of models trained with and without the early-stop strategy, by comparing performance on multiple testset, as shown in Tab.~\ref{tab:trainingsetting}. We denote models trained with and without the strategy as `unsupervised-w crit.' and `unsupervised-w/o crit.' separately. For both models trained with WHU-Stereo-train and US3D-train, early-stop strategy helps to gain better and stable generalization ability on testsets. On all the five testsets, as shown in Fig.~\ref{fig:trainingsetting}, test results of models trained with early-stop strategy show sharper edges of buildings, less noise and disturbance in large untextured areas, and less missing details of small buildings. It can be concluded that the use of the early-stop consistency criterion can lead to a superior and more stable generalization effect on various different datasets, without over-adaption of one certain dataset.
\newpage
\section{Conclusion}
\vspace{-12pt}
\label{sec:conculsion}
In our comprehensive analysis of the key factors influencing the generalization performance of remote sensing stereo matching task, the following insights emerge: For selection of train set, it is imperative to utilize data that exhibits a similar regional target distribution to the test domain, while simply increasing amount of train set or employing data obtained by the same sensors cannot help. For model structure, cascaded structure that adeptly accommodates multi-scale features is effective for generalization. Furthermore, the utilization of pre-trained models also significantly contributes to improving the overall performance. For training manner, given the impractically of acquiring data from exact same domain to train, unsupervised methods generalize better and are advised to be considered as priority. For future work, there lacks a quantitative definition of regional target distribution similarity, which should be addressed as a promising research topic contributing to advancements in the geoscience field. 

\bibliographystyle{IEEEtran}
\bibliography{refs}

\end{document}